\documentclass[a4paper,fleqn]{cas-dc}

\usepackage[numbers]{natbib}
\usepackage[dvipsnames]{xcolor}

\begin{document}
\let\WriteBookmarks\relax
\def\floatpagepagefraction{1}
\def\textpagefraction{.001}
\definecolor{forestgreen}{RGB}{50, 150, 30}

\shorttitle{Federated Scene Graph Generation}

\shortauthors{S. Ha et al.}  

\title [mode = title]{Benchmarking Federated Learning for Semantic Datasets: Federated Scene Graph Generation}       

\author[1]{SeungBum Ha}[orcid=0009-0000-1621-3551]
\cormark[1]
\ead{ethereal0507@unist.ac.kr}

\author[1]{Taehwan Lee}
 [orcid=0009-0001-4745-6051]
\cormark[1]
\ead{taehwan@unist.ac.kr}

\author[3]{Jiyoun Lim}
\ead{kusses@etri.re.kr}

\author[1,2]{Sung Whan Yoon}
 [orcid=0000-0002-7202-2837]
\cormark[2]
\ead{shyoon8@unist.ac.kr}

\affiliation[1]{
organization={Graduate School of Artificial Intelligence, Ulsan National Institute of Science and Technology},
city={Ulsan},
country={Korea}}

\affiliation[2]{
organization={Department of Electrical Engineering, Ulsan National Institute of Science and Technology},
city={Ulsan},
country={Korea}}

\affiliation[3]{
organization={Electronics and Telecommunications Research Institute (ETRI)},
city={Daejeon},
country={Korea}}

\cortext[co]{Co-first author}
\cortext[cor1]{Corresponding author}

\begin{abstract}
Federated learning (FL) enables decentralized training while preserving data privacy, yet existing FL benchmarks address relatively simple classification tasks, where each sample is annotated with a one-hot label.
However, little attention has been paid to demonstrating an FL benchmark that handles complicated semantics, where each sample encompasses diverse semantic information, such as relations between objects.
 Because the existing benchmarks are designed to distribute data in a narrow view of a single semantic, managing the complicated \textit{semantic heterogeneity} across clients when formalizing FL benchmarks is non-trivial.
In this paper, we propose a benchmark process to establish an FL benchmark with controllable semantic heterogeneity across clients: two key steps are (i) data clustering with semantics and (ii) data distributing via controllable semantic heterogeneity across clients.
As a proof of concept, we construct a federated PSG benchmark, demonstrating the efficacy of the existing PSG methods in an FL setting with controllable semantic heterogeneity of scene graphs. We also present the effectiveness of our benchmark by applying robust federated learning algorithms to data heterogeneity to show increased performance. To our knowledge, this is the first benchmark framework that enables federated learning and its evaluation for multi-semantic vision tasks under the controlled semantic heterogeneity. Our code is available at \href{https://github.com/Seung-B/FL-PSG}{https://github.com/Seung-B/FL-PSG}.
\end{abstract}

% Research highlights
%\begin{highlights}
%\item We present the FL framework for the Panoptic Scene Graph Generation Task.
%\item We suggest the method to construct the heterogeneity for multi-semantic datasets.
%\item By evaluating the Panoptic Scene Graph Generation algorithms in Federated Learning, we analyze the effect of heterogeneity.
%\end{highlights}

% Keywords

\maketitle

% Main text
\section{Introduction}

Federated learning (FL) has drawn considerable attention as a key framework to enable decentralized deep model training from private data of numerous clients.
The FL framework communicates the model parameters between the clients and the
server; to keep the distributed local data private, the server cannot access
data samples of clients~\cite{fedavg}.
The property that FL preserves data privacy makes it more crucial when deep models handle license- or privacy-sensitive data, e.g.,~clinical data from medical institutions, licensed content from providers, and broadcasting stations.

Along with the rapid algorithmic development of FL, significant efforts have been dedicated to constructing FL benchmarks that enable reliable and rigorous evaluations of methods.
The existing FL benchmarks mostly rely on the existing datasets, such as
CIFAR~\cite{CIFAR} and Twitter~\cite{leaf'18}, etc. Therefore researchers focus
on devising a decentralized training setting with controllable factors, such as
data heterogeneity across clients, number of clients, and participation ratio.
Among the factors of FL settings, \textit{data heterogeneity} works as the most
crucial factor that exhibits the efficacy of different FL algorithms; when the
data distribution strongly deviates across clients, a federation of local
models typically fail with drastic performance drops~\cite{li2019convergence}.
Researchers have mainly focused on data heterogeneity constructed by a single
label, which is straightforward. Therefore, prior works diversify the
distribution across clients via Dirichlet distribution~\cite{FedDyn'21} or
shard- or chunk-wise assignment of data~\cite{fedavg}.

Herein, we point out two key limitations of the existing FL benchmarks.
Firstly, the current benchmarks mainly handle classification or regression tasks, where each sample consists of a single label. 
However, deep learning tasks are becoming far beyond classification or recognition, and complicated jobs are being considered to understand in-depth semantic information. Therefore, extending the current FL benchmark process to complicated semantics is necessary.

\begin{figure*}[t!]
\centerline{\includegraphics[width=\linewidth]{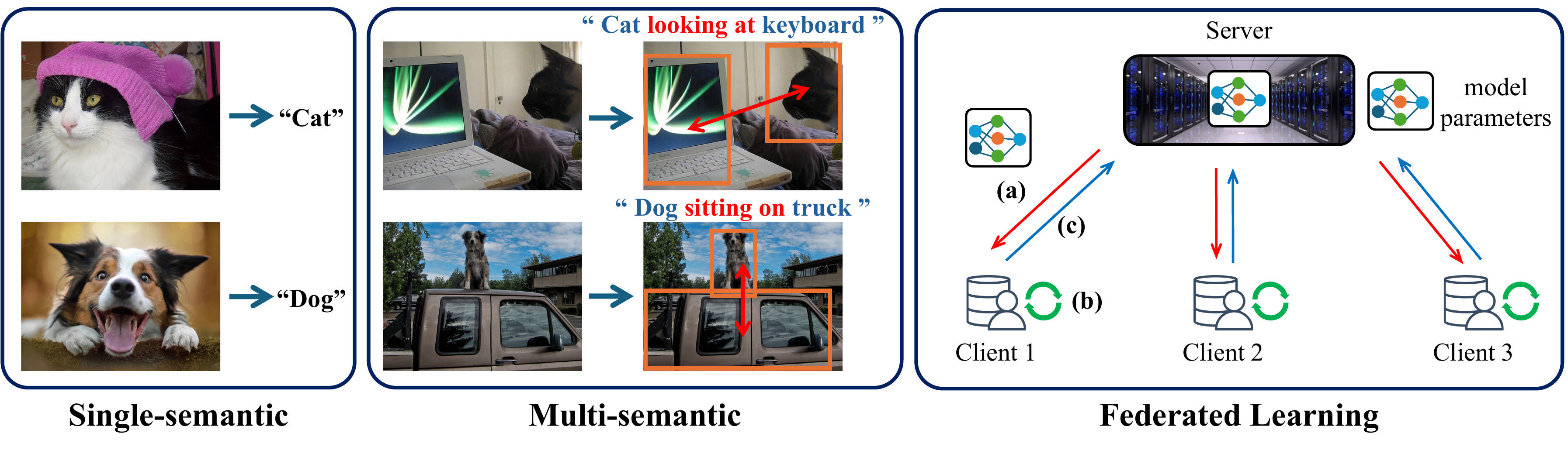}}
\caption{
Examples of single-semantic (Left: classification) and multi-semantic (Middle: scene graph generation) tasks, and the 
overview of the federated learning process (Right).
}
\label{Example_Semantic}
\end{figure*}

Second, there does not exist a task-agnostic FL benchmark process that devises controllable \textit{semantic heterogeneity}.
 The classification task of FL focuses on a single semantic (the class label) in the left of  \ref{Example_Semantic} and constructs heterogeneity based on the label. 
 In contrast, devising heterogeneity is non-trivial when each sample contains multiple semantics.
As shown in the middle of  \ref{Example_Semantic}, Scene Graph Generation, which understands the complicated semantics of an image, a single image contains multiple objects (`cat', `keyboard', `dog' and `truck'), predicates (`looking at' and `sitting on') and relations (`cat' $\rightarrow$ `keyboard', `dog' $\rightarrow$ `truck'). 
It remains unexplored to construct the heterogeneity with controllable and complex semantics.

In this study, we propose a comprehensive FL benchmark process for evaluating FL algorithms on multi-semantic datasets while controlling semantic heterogeneity. 
To break the limitations of existing studies, our process encompasses two key steps: (i) discovering the semantic clusters by utilizing the collection of multiple annotations. and (ii) distributing data samples to multiple clients by considering the heterogeneity. 
The simulation results reveal that the methods tailored to tackle the long-tailed problem in the Panoptic Scene Graph Generation (PSG) task, where some objects and predicates are more dominant than others, are robust in handling semantic heterogeneity in FL.

\section{Related Works}
FL has emerged as a framework for training deep learning models in a decentralized setting, enabling the preservation of data privacy for clients.
 We briefly introduce preliminaries by focusing on the foundational baseline,
i.e.,~FedAvg,~\cite{fedavg}.
FL setting contains the single server and $K$ clients. 
The training process proceeds iteratively in rounds. Each round follows the three-step procedure shown on the right of  \ref{Example_Semantic}, summarized as follows:
 \textbf{(a)}: The server distributes a global model to clients.
\textbf{(b)}: Each client initializes the local model with the distributed one and trains a neural network using their dataset
\textbf{(c)}: Clients upload the locally trained model parameters to the server. Then, the server averages the aggregated models and performs the step \textbf{(a)} again. 
The aggregation step is represented as $w^{t+1}=\sum^{S_t}_{k=1} \frac{n_k}{n}w^t_k$, 
   where $n_k$ is the number of data samples on client $k$, $S_t$ is the set of selected clients at round $t$, and $n$ is the total number of data samples across $S_t$ clients. When $S_t$ equals $K$, all the clients participate in the aggregation step, and in the case of $S_t<K$, partial clients participate in the aggregation step.

In FL systems, each user has independent data with different distributions, which can cause the trained local models to diverge, negatively impacting the performance and convergence of FL.
Therefore, researchers have mainly been dedicated to handling the case with a substantial heterogeneity of data across clients, resulting in diverse strategies. 
For example, FedAvgM~\cite{fedavgm} leverages global-model updates as momentum.
At each round, the server keeps a momentum vector formed by accumulating the
difference between consecutive global models. FedAdam~\cite{fedopt} treats
update differences as gradients optimized by Adam. Recently,
regularization-based methods such as  $\ell_1$-Fed~\cite{shi2023federated},
which alleviates similar distribution through the addition of an $\ell_1$
sparsity constraint to the global model, have also been reported. And
FedGF~\cite{fedgf'24}, which resolves the difference between the local and
global objectives by raising the generalization ability with Sharpness-Aware
Minimization~\cite{SAM'20}.

 When we construct a heterogeneous distribution with a simple semantics, such
as a single target label, unified strategies exist to impose heterogeneity
across clients by diversifying the prior distribution of the target label~\cite{noniidsurvey}.
Specifically, two main strategies include \textbf{(i)} sampling the prior
distribution of each client from Dirichlet distribution~\cite{FedDyn'21}, and
\textbf{(ii)} chunking per-class data samples into multiple shards, where a
fixed number of shards is allocated to each client, resulting in heterogeneity
between clients~\cite{achituve2021personalized,lim2024metavers}.

\subsection{Panoptic Scene Graph Generation (PSG)}

Scene graphs are crucial for scene understanding in computer vision tasks,
representing objects (nodes), denoted by bounding boxes or pixel-wise
segmentation, and predicates (relationships, edges) in a graph
structure~\cite{SGG'PRL, vg, lin2014microsoft}.
Predicting the bounding boxes and relationships between bounding boxes constitutes scene graph generation. 
The PSG task has been proposed by~\cite{yang2022panoptic,ZHAO202556}, delving
deeper into scene graph generation using panoptic segmentation masks instead of
bounding boxes. The difference between PSG and classic scene graph generation
is that PSG uses panoptic segmentation~\cite{kirillov2019panoptic} masks rather
than bounding boxes. 
Moreover, the scene graph generation tasks face
the long-tailed problem~\cite{desai2021learning,Longtail-FL'PRL}. Positional
relationships among objects constitute the majority of the predicates, leading
to a visual relationship complexity of $\mathcal{O}(N^2R)$ for $N$ objects and
$R$ predicates~\cite{chang2021scene}. This exacerbates the long-tailed
problem in SGG datasets, prompting various approaches, such as utilizing
self-attention to characterize complex interactions, thereby facilitating the
understanding of object and relation semantics~\cite{SGG-Attention'PRL}, and
estimating confidence scores and weighting high-uncertainty cases more heavily
during training~\cite{SGG-uncertainty'PRL}. Also, the self-supervised local
pseudo-attribute is utilized to reinforce tail-class
representations~\cite{kim2023local}.

To perform challenging vision tasks such as PSG, having more data typically leads to training better models. However, in reality, the photos held by different clients are unlikely to be similar, and collecting such data on the server for training poses a threat to data privacy. 
Despite this, no attempts have been made to apply FL to the PSG task,
underscoring the necessity of this research. Moreover, the data heterogeneity
issue in FL closely resembles the well-known long-tailed problem in scene graph
generation tasks and real-world data~\cite{Longtail-FL'PRL}.

\section{A Benchmark Process for FL with Multi-Semantic Datasets}\label{FL-PSG Dataset}

For a given multi-semantic dataset, each data sample contains multiple annotations, i.e.,~$(x,\mathcal{Y})\in\mathcal{D}$, where $x$ is an input, $\mathcal{Y}=\{y_1, \ldots,y_M\}$ is a set of multi-semantic labels, $M$ is the possible number of labels for each data sample, and $\mathcal{D}$ represents the dataset.
Here, we introduce our benchmark process to distribute the multi-semantic data samples to clients with controllable semantic heterogeneity.
The key steps are twofold: (i) discovering data clusters with different semantics and (ii) data partitioning with controllable semantic heterogeneity across clients.

\subsection{Discovering Data Clusters: $K$-means Clustering of Category Tensor}
\label{sec:Clustering}

\begin{figure*}[th!]
	\centering
         \includegraphics[width=1\linewidth]{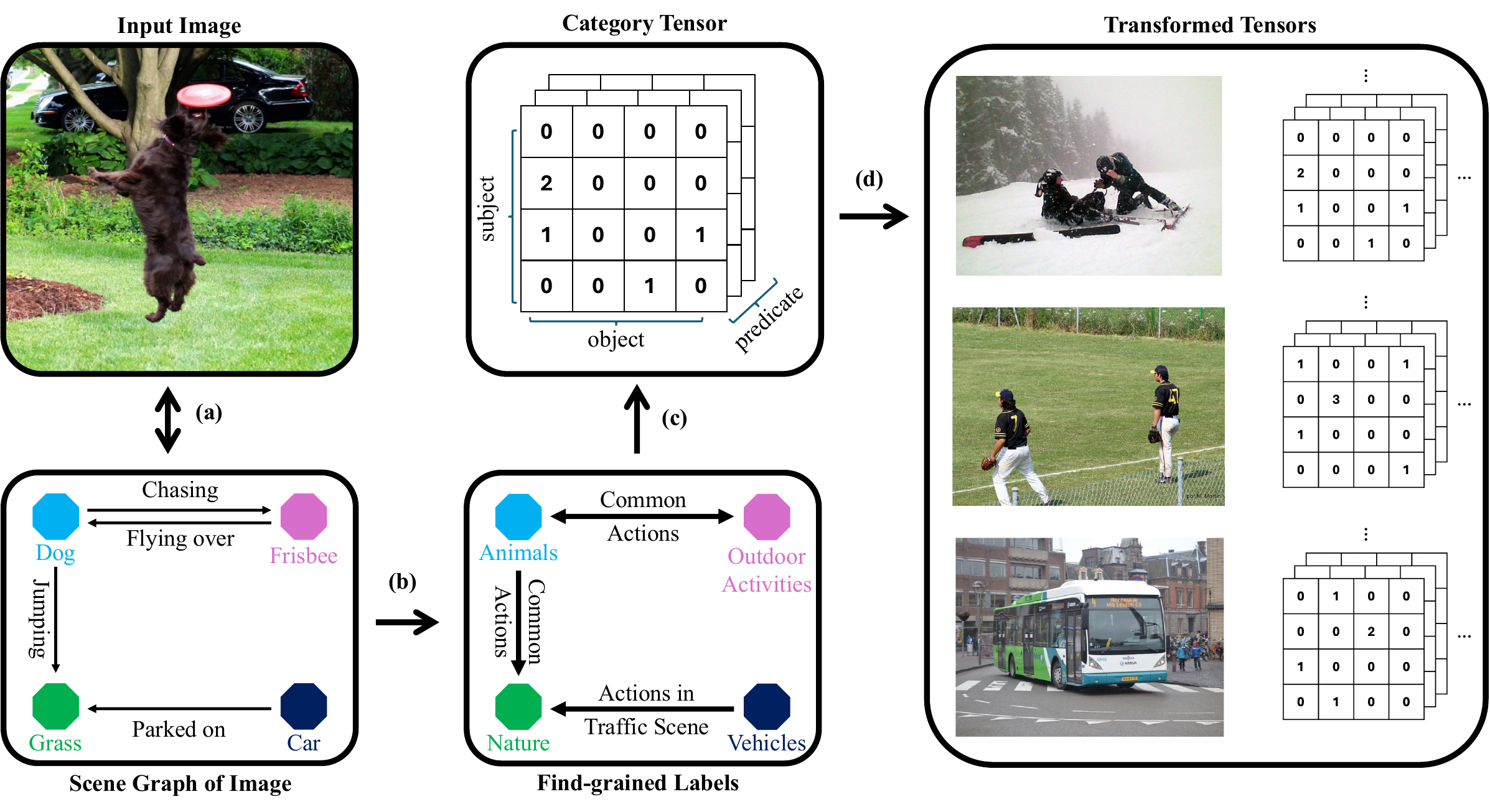}
	\caption{\textbf{Category Tensor K-means Clustering Pipeline}. 
    \textbf{(a)} Construct a graph using the objects, subjects, and predicates of each image.
    \textbf{(b)} Map each label into the super-classes of fine-grained labels.
    \textbf{(c)} Convert categorized relations (subject, object, and predicate) into the category tensor.
    \textbf{(d)} Transform every input image into a category tensor and perform K-Means Clustering.
    }\label{Fig:Image_Category_Matrix}
\end{figure*}

For a given multi-semantic $\mathcal{Y}$, we transform it to \textit{category tensor} $\mathcal{F}$ by allocating each label $y_i$ into an orthogonal axis of the tensor, i.e.,~$\mathcal{F}(\mathcal{Y})\in \mathbb{R}^{N_1\times \cdots \times N_L}$, where there are $N_1, \ldots, N_L$ possible categories for each respective label of $\mathcal{Y}$. 
We then apply $K$-means Clustering on the collection of 
$\mathcal{F(Y)}_{1}^{|\mathcal{D}|}$ of overall dataset:
\begin{equation} \label{eq:cluster}
\mathcal{K}\Big( \mathcal{F(Y)}_{1}^{|\mathcal{D}|} \Big) \rightarrow \{\mathcal{C}_1, \cdots \mathcal{C}_n\},
\end{equation}
where $n$ is the number of clusters that can be determined depending on the dataset and $\mathcal{C}_i$ indicates the collection of samples assigned to $i$th cluster.
With the obtained clustering, we can transform each data sample $(x,\mathcal{Y})$ into $(x,\mathcal{C}_i)$ to impose the cluster label with semantic information, which is a one-hot label with $1$ for the assigned cluster.
As a result of the clustering process, we can perform the label-based partition while fully utilizing the multi-semantic information of each data sample with its corresponding cluster label $\mathcal{C}$. We present the overall pipeline to  \ref{Fig:Image_Category_Matrix}.

\subsection{Data Partition with Semantic Heterogeneity} 
\label{sec:data-partition}

We acquire $n$ clusters from Eq.~\eqref{eq:cluster}. 
 It trivially raises the issue that the clusters are not evenly distributed, so the number of samples assigned to each cluster would deviate for different clusters, i.e.,~\textit{cluster imbalance}.
The \textit{cluster imbalance} prevents rigorous evaluations of FL models to handle semantic heterogeneity because a model becomes overfitted to dominant clusters without balanced training across different semantics.
The cluster imbalance stems from the long-tailed problem, a key challenge in scene graph generation datasets. In other words, we have to create data heterogeneity for FL, which further complicates distinguishing it from the long-tailed problem. If the amount of data in each cluster is equalized, the long-tailed problem can be effectively alleviated.
Furthermore, considering the FL scenario, this cluster imbalance will likely bias the update of the global model in the direction of users belonging to the dominant cluster. It causes overfitting to a dominant cluster, which makes it difficult to fairly compare each method closely.
Consequently, we need to equalize the data quantity of each cluster: $\hat{\mathcal{C}}_k = \textnormal{Sample}(\mathcal{C}_k, m)$, for all $1\leq k \leq n$, where $m = \min_{k\in [n]}\{| \mathcal{C}_k| \}$, $| \mathcal{C}_k| $ is the cardinality of the $k$th cluster $\mathcal{C}_k$, and $\textnormal{Sample}(\mathcal{C}_k, m)$ functions to randomly select $m$ data samples from cluster $\mathcal{C}_k$.
 We apply the label-based partition based on these clusters to impose semantic heterogeneity. 
Our benchmark suggests two partition strategies as follows.

\textbf{Shard-based partition:}
Each client chooses $p(\leq n)$ clusters. 
We then split each cluster into disjoint shards or chunks, where the number of shards equals the number of clients who selected the cluster. After splitting, the shards are distributed to the corresponding clients.
If $p=n$, all clients are assigned to all clusters, making the data distribution homogeneous. The distribution is close to heterogeneous for smaller $p$ values.

\textbf{Dirichlet distribution-based partition:}
From the strategy suggested in~\cite{FedDyn'21}, the amount of data each client
takes from cluster $k$ is governed by the sampling from the Dirichlet
distribution. We design the non-IID data partition into $U$ clients by
sampling a multinomial probability vector for each client $ u $, denoted
as:
$\mathbf{p}_u \sim \mathrm{Dir}_n(\boldsymbol{\alpha})$
where the probability vector is $ \mathbf{p}_u = (p_{u,1}, p_{u,2}, \ldots , p_{u,n})$, 
  and the concentration parameter of Dirichlet distribution is $
\boldsymbol{\alpha} = (\alpha_1, \alpha_2, \ldots , \alpha_n)$, 
for $\alpha_i > 0,  \forall i \in \{1,2, \ldots ,n\}$.
 Each $u$th client samples training data from a dataset according to the proportion $p_{u,i}$ without replacement for each cluster $i$. 
The data heterogeneity is controlled by $\boldsymbol{\alpha}$:
As $\boldsymbol\alpha \to \infty$, the data distribution goes to IID; in contrast, as $\alpha \to 0$, it goes to non-IID.

\subsection{Proof-of-concept: FL Benchmark for Panoptic Scene Graph Generation (PSG)}

We provide a proof-of-concept of our FL benchmark process by constructing the FL benchmark for PSG dataset.

\textbf{(i) Discovering data clusters:} PSG dataset contains object, subject, and predicate labels for each image sample. For simplicity, we utilize 13 object/subject categories and 7 predicate categories, which are the super-classes of fine-grained labels
 Therefore, the dimension of the category tensor is $\mathcal{F}(\mathcal{Y})\in\mathbb{R}^{13\times 13\times 7}$.
We perform $K$-means Clustering for the category tensor to obtain multiple semantic clusters, obtaining 5 clusters with discriminated semantics.\footnote{We attach the detailed description in  Appendix A.}:
\begin{equation}
 \label{eq:cluster-PSG}
\mathcal{K}\Bigl( \mathcal{F(Y)}_{1}^{| \mathcal{D}| } \Bigr) \rightarrow \{\mathcal{C}_1, \mathcal{C}_2, \mathcal{C}_3, \mathcal{C}_4, \mathcal{C}_5\}.
\end{equation}

\textbf{(ii) Data partition:} Based on the discovered 5 semantic clusters, our benchmark provides two options for data distribution: (i) Shard-based partitioning and (ii) Dirichlet distribution-based partitioning.
As partitioning becomes heterogeneous, the data distribution at clients strongly deviates in the sense of semantic clusters.
 Otherwise, the data distribution of clients becomes homogeneous, yielding evenly distributed semantic information.

\section{Experiments: Benchmarks for PSG in FL}
\label{sec:exp}

\subsection{Experiment Settings}

We evaluate the existing panoptic scene graph generation (PSG) models on our
benchmark with the following methods: IMP~\cite{imp}, MOTIFS~\cite{motifs},
VCTree~\cite{vctree}, and GPS-Net~\cite{gpsnet}.
Because these PSG methods use the same pretrained object detector Faster
R-CNN~\cite{fastrcnn}, and the communication cost is crucial in FL scenarios,
we freeze the pretrained object detector and focus on predicate classification.
Therefore, each client trains and aggregates the relation head, responsible for
processing predicates by capturing the semantic relationships between objects.

An imbalanced dataset contains a number of data points in each cluster that is not equal after clustering. We randomly sampled the data from each cluster to match the quantity of the smallest cluster to eliminate cluster imbalance. This process ensured that all clusters had the same data (2.2K images), resulting in a balanced dataset.

\textbf{Experiment setups:}
We set up an FL scenario with one server and 100 clients, distributing the
training data of the existing PSG dataset~\cite{yang2022panoptic} to the 100
clients. The test data for our benchmark is the same as the PSG test dataset. 
Five active clients are randomly selected in each round, and the test data is evaluated using the aggregated global model from the server.
Each client performs local training with one epoch and a batch size 16. The total number of training rounds is 100, and we report the R/mR@K performance of the final averaged model.
Following the benchmark in~\cite{yang2022panoptic}, we set the SGD optimizer to
a local optimizer with a learning rate of 0.02, momentum of 0.9, weight decay
of 0.0001, and gradient clipping with a max L2 norm of 35.

\textbf{Clustered PSG Dataset Description:} By examining the samples for each cluster, we observe the following features for each cluster and the imbalance between clusters\footnote{The visualization of the clustering is in Appendix \ref{Clustered PSG Dataset}}:

\begin{itemize}

    \item \textbf{Cluster 1} (occupying 5\% of datasets)

 We observe that it contains a large number of \textbf{animal objects} compared to others. The predicates are composed of actions that animals trivially perform.

    \item \textbf{Cluster 2} (occupying 58\% of datasets)

 This cluster is dominated by \textbf{daily photographs of people}, which constitutes the largest portion of PSG dataset. This cluster is mainly related to daily activities by human beings that frequently appear in daily life. 

    \item \textbf{Cluster 3} (occupying 11\% of datasets)

 This cluster mainly includes urban landscape and transportation photos, which encompass many predicates related to vehicles, such as `parking on' and `driving (on).'

    \item \textbf{Cluster 4} (occupying 7\% of datasets)

 This cluster is composed of \textbf{sports} and \textbf{kinetic} images, containing predicates such as `playing,' which are more prevalent than others.

    \item \textbf{Cluster 5} (occupying 19\% of datasets)

 This cluster corresponds to \textbf{urban/nature-combined landscapes}, which typically include buildings, the sky, and a river in the images. Due to objects related to natural elements, the predicates in this cluster are predominantly positional rather than action-oriented.

\end{itemize}
Notably, Clusters 2 and 4 contain somewhat similar images, mainly of `people'. However, the predicates in Cluster 4 relate to sports, clearly distinguishing it from Cluster 2. Also, Cluster 3 and 5 look similar because of urban landscapes, but Cluster 3 tends to focus on cityscapes with transportation, and Cluster 5 focuses on urban/nature-combined views.

\begin{table*}[t!]
\small 
\centering
\caption{Comparison of the performances of PSG methods on the proposed FL benchmark}
\label{main-table}
\resizebox{0.97\textwidth}{!}
{\begin{tabular}{ccccccccc}
\Xhline{3\arrayrulewidth}
\multirow{2}{*}{\shortstack{R/mR\\@K}}                        & \multirow{2}{*}{Method} & \multirow{2}{*}{CL$^\dagger$}  & \multirow{2}{*}{Random}& \multicolumn{2}{c}{Shard}   & \multicolumn{3}{c}{Dirichlet distribution}               \\
                                                              &                             &                        &                         & IID          & non-IID      & $\alpha=10($$\approx$ IID) & $\alpha=1$   & $\alpha=0.2$ \\
\Xhline{1.5\arrayrulewidth}
\multicolumn{1}{c|}{\multirow{4}{*}{\shortstack{R/mR\\@20}}}  & IMP                         & 16.54 / 6.55          & 12.45 / 3.08            & 12.62 / 3.20 & 11.26 / 2.28 & 12.31 / 3.36              & 12.10 / 2.92 & 9.31 / 1.78 \\
\multicolumn{1}{c|}{}                                         & MOTIFS                      & \underline{16.97} / \underline{7.56}          & \underline{13.54} / \underline{4.60}            & \underline{13.26} / \underline{4.64} & \underline{13.33} / \underline{4.06} & \underline{13.33} / 4.39              & \underline{13.34} / 4.09 & \underline{13.25} / \underline{4.28} \\
\multicolumn{1}{c|}{}                                         & VCTree                      & 16.80 / 7.20          & 12.73 / 4.38            & 13.00 / 4.57 & 12.49 / 3.99 & 13.00 / \underline{4.42}              & 12.86 / \underline{4.36} & 13.06 / 4.17 \\
\multicolumn{1}{c|}{}                                         & GPS-Net                     & \textbf{18.00} / \textbf{7.83}          & \textbf{13.93} / \textbf{5.98}            & \textbf{14.83} / \textbf{6.90} & \textbf{14.57} / \textbf{5.90} & \textbf{14.88} /\textbf{ 6.33}              & \textbf{14.82} / \textbf{6.16} & \textbf{14.38} /\textbf{ 5.91} \\
\Xhline{1.5\arrayrulewidth}
\multicolumn{1}{c|}{\multirow{4}{*}{\shortstack{R/mR\\@50}}}  & IMP                         & 17.87 / 6.96          & 13.89 / 3.44            & 13.97 / 3.53 & 12.57 / 2.59 & 13.79 / 3.73              & 13.40 / 3.23 & 10.83 / 2.03 \\
\multicolumn{1}{c|}{}                                         & MOTIFS                      & \underline{18.59} / \underline{8.01}          & \underline{15.07} / \underline{5.05}            & \underline{14.82} / \underline{5.06} & \underline{14.92} / \underline{4.48} & \underline{14.77} / 4.71              & \underline{14.63} / 4.44 & \underline{14.77} / \underline{4.64} \\
\multicolumn{1}{c|}{}                                         & VCTree                      & 18.54 / 7.70          & 14.20 / 4.75            & 14.50 / 4.94 & 14.04 / 4.41 & 14.32 / \underline{4.82}              & 14.34 / \underline{4.78} & 14.51 / 4.56 \\
\multicolumn{1}{c|}{}                                         & GPS-Net                     & \textbf{19.69} /\textbf{ 8.30}          & \textbf{15.63} /\textbf{ 6.51}            & \textbf{16.42} / \textbf{7.37} & \textbf{16.37} / \textbf{6.36} & \textbf{16.46} / \textbf{6.74}              & \textbf{16.34} / \textbf{6.62} & \textbf{16.01} / \textbf{6.36} \\
\Xhline{1.5\arrayrulewidth}
\multicolumn{1}{c|}{\multirow{4}{*}{\shortstack{R/mR\\@100}}} & IMP                         & 18.37 / 7.11          & 14.46 / 3.56            & 14.45 / 3.65 & 13.06 / 2.68 & 14.48 / 3.89              & 13.92 / 3.35 & 11.25 / 2.10 \\
\multicolumn{1}{c|}{}                                         & MOTIFS                      & \underline{19.15} / \underline{8.14}          & \underline{15.64} / \underline{5.16}            & \underline{15.38} / \underline{5.20} & \underline{15.43} / \underline{4.65} & \underline{15.33} / 4.86              & \underline{15.15} / 4.62 & \underline{15.18} / \underline{4.71} \\
\multicolumn{1}{c|}{}                                         & VCTree                      & 19.02 / 7.82          & 14.69 / 4.87            & 14.97 / 5.05 & 14.62 / 4.54 & 14.87 / \underline{4.97}              & 14.90 / \underline{4.90} & 15.03 / 4.68 \\
\multicolumn{1}{c|}{}                                         & GPS-Net                     & \textbf{20.28} / \textbf{8.47}          & \textbf{16.34} / \textbf{6.66}            & \textbf{17.08} / \textbf{7.55} & \textbf{16.91} / \textbf{6.49} & \textbf{17.10} / \textbf{6.91}              & \textbf{16.84} / \textbf{6.77} & \textbf{16.55} / \textbf{6.51} \\
\Xhline{3\arrayrulewidth}
\end{tabular}}
\footnotesize{\\ $^\dagger$ For centralized learning (CL) is with a centralized dataset without considering the FL settings.\\ \textbf{Bold} refers the best performance and \underline{underline} denotes the $2^{\text{nd}}$ performance.}
\end{table*}

\textbf{Benchmark setups:}
We randomly sampled data from each cluster to ensure an equal amount of data for each cluster to ease the cluster imbalance. We test 6 types of data partitioning as follows: 

\textbf{(1) Random:} Data is distributed randomly among all clients, ensuring nearly equal sizes for each. 

\textbf{(2) Shard-based partition IID:} We set $p=5$, where $p$ is the number of clusters that client sample from. When $p$ equals the number of clusters, the data from each cluster is equally distributed among 100 clients.

\textbf{(3) Shard-based partition non-IID:} We set $p=1$ for imposing semantic heterogeneity. Each cluster is assigned 20 clients, and all clients have the same amount of data. 

\textbf{(4), (5) and (6) Dirichlet distribution-based partition:} 
We test three levels of semantic heterogeneity by using $\alpha = [10, 1, 0.2]$ to simulate from an IID to a non-IID case.

\textbf{Metrics:}
By following~\cite{yang2022panoptic} which suggested the PSG task, we use
`Recall@K (R@K)' and `mean Recall@K (mR@K)' as the performance metrics, which
calculate the triplet recall and mean recall for every predicate category,
given the top $K\in[20,100]$ triplets from a PSG method. 
 Moreover, R@K is dominated by high-frequency relations, and mR@K assigns equal weight to all relation classes. In datasets with severe long-tailed problems, e.g.,~PSG dataset, mR@K can provide more meaningful insights into model performance.

\subsection{In-depth Analysis}

Our intuition is that the performance of models is expected to show the following order: Centralized learning(CL) c$\geq$ IID $\geq$ Random $\geq$ non-IID, when our benchmark effectively imposes semantic heterogeneity in the FL setting. The experimental results also follow our intuition and validate the effectiveness of our benchmark.

\textbf{Results:}
\tabref{main-table} shows the test accuracy on the test set of the PSG dataset.\footnote{In  Appendix C we attach additional experiments including Convergence behavior, Communication cost, Cluster Imbalance effect, Extension to FL algorithms, and various FL scenarios.}
We have focused on the Mean Recall (`mR') performance. Also, we focus on the most challenging case with $K=20$.

\textbf{(i) CL vs. IID.} The performance has been mostly degraded when comparing CL and IID cases.
The averaged gaps for mR@20 are $-2.45\%$ and $-2.71\%$, for `Shard-IID' and `Dir($\alpha=10$)'. Each client has approximately 114 images, and due to the limited data, there appears to be a performance difference between the CL and IID scenarios.
CL can collectively form a mini-batch across clients, but IID forms a mini-batch per client in a decentralized manner.

\textbf{(ii) IID vs. Random.} When data is randomly divided, it will tend to have a distribution close to IID so that there is a minimal performance drop. 
The averaged gaps for mR@20 are $-0.32\%$ and $-0.12\%$, for `Shard-IID' and `Dir($\alpha=10$)', respectively. 
The results confirm that the random partitioning naively conducted in prior studies is unsuitable for imposing semantic heterogeneity, showing similar results as the IID case.

\textbf{(iii) IID vs. non-ID.} We confirm large performance degradations in most cases.
First, in the case of a shard-based partition, the averaged gap for mR@20 is $-0.77\%$. Second, in the case of the Dirichlet distribution-based partition, i.e.,~comparing Dir($\alpha=10$) and Dir($\alpha=0.2$), the averaged gap for mR@20 is shown to be $-0.64\%$.
The performance drops from IID to non-IID reveal that PSG methods struggle to aggregate a global model under strong semantic heterogeneity. 
 MOTIFS shows the outliers in mR, where the moderate non-IID case ($\alpha=1$) compared to the non-IID case ($\alpha=0.2$) shows minimal differences: 4.09\% vs. 4.28\% in mR@20, and 4.62\% vs. 4.71\% in mR@100.
It looks unexpected, but it is not a considerable amount. Also, we want to point out that when $\alpha=10$, which is the IID case, the performance becomes maximized: 4.39\% in mR@20 and 4.86\% in mR@100, which coincides with our expectations. 
We conjecture that the behavior at the moderate non-IID can be a little shaky in a few cases, but it finally behaves as expected in the IID case.
Although the results may seem unexpected, the differences are not significant. Notably, when $\alpha=10$, corresponding to the IID case, shows the best performance: 4.39\% in mR@20 and 4.86\% in mR@100, aligning with our expectations.
Based on this observation, we conjecture that the behavior at the moderate non-IID can be a little shaky in a few cases, but it behaves as expected in the IID case.

\textbf{PSG Model comparisons:}
We discuss the robustness of the existing PSG methods against semantic heterogeneity. We conclude that IMP is relatively vulnerable in handling semantic heterogeneity in FL, i.e.,~a large gap of $-1.58\%$ for mR@20 is observed when comparing Dir($\alpha=10$) and Dir($\alpha=0.2$).
It has a smaller model architecture and suffers from the long-tailed problem in the PSG dataset. We conjecture that the aspects of IMP lead to notable performance drops in our non-IID testing.
VCTree includes a tree construction process trained via reinforcement learning, resulting in a more complex model structure than MOTIFS. Consequently, in the FL scenario with small-scale client data, its performance is degraded.
Because GPS-Net employs key elements, e.g.,~DMP, NPS-loss, and ARM, to resolve the long-tailed problem, 
we conjecture that it yields the outperforming results of GPS-Net in our FL benchmarks.

\subsection{Extension to FL algorithms}\label{Extension to FL algorithms}

\begin{table*}[tb!]
\centering
\caption{Comparison of the FedAvg, FedAvgM, FedAdam performances of PSG methods.}
\label{FedAvgM-Result}
{%
\begin{tabular}{ccccc}
\Xhline{3\arrayrulewidth}
\multirow{2}{*}{R/mR@K} & \multirow{2}{*}{Method} & \multicolumn{1}{c}{FedAvg} & \multicolumn{1}{c}{FedAvgM\cite{fedavgm}} & \multicolumn{1}{c}{FedAdam\cite{fedopt}} \\
 &  & Shard non-IID & Shard non-IID & Shard non-IID \\ \hline
\multicolumn{1}{c|}{\multirow{4}{*}{R/mR@20}} 
                        & IMP & 11.26 / 2.28 & 13.23 / 3.83 \textcolor{forestgreen}{(+1.55\%)} & 13.32 / 4.78 \textcolor{forestgreen}{(\textbf{+2.50\%})}
                        \\
\multicolumn{1}{c|}{} & MOTIFS & \underline{13.33} / \underline{4.06} & \underline{15.47} / \underline{5.80} \textcolor{forestgreen}{(\textbf{+1.74\%})} &
                        \textbf{15.89} / \underline{5.56} \textcolor{forestgreen}{(+1.50\%)}
                        \\
\multicolumn{1}{c|}{} & VCTree & 12.49 / 3.99 & 15.39 / 5.66 \textcolor{forestgreen}{(+1.67\%)} &
                         15.53 / 5.09 \textcolor{forestgreen}{(+1.10\%)}
                        \\
\multicolumn{1}{c|}{} & GPS-Net & \textbf{14.57} / \textbf{5.90} & \textbf{16.18} / \textbf{5.91} \textcolor{forestgreen}{(+0.01\%)} &
                        \underline{15.66} / \textbf{5.98} \textcolor{forestgreen}{(+0.08\%)}
                        \\ \hline
\multicolumn{1}{c|}{\multirow{4}{*}{R/mR@50}} 
                        & IMP & 12.57 / 2.59 & 14.73 / 4.24 \textcolor{forestgreen}{(+1.65\%)}&
                        15.03 / 5.41 \textcolor{forestgreen}{(\textbf{+2.82\%})}
                        \\
\multicolumn{1}{c|}{} & MOTIFS & \underline{14.92} / \underline{4.48} & \underline{17.23} / \underline{6.23} \textcolor{forestgreen}{(\textbf{+1.75\%})} &
                        \textbf{17.66} / \underline{6.01} \textcolor{forestgreen}{(+1.53\%)}
                        \\
\multicolumn{1}{c|}{} & VCTree & 14.04 / 4.41 & 17.02 / 6.10 \textcolor{forestgreen}{(+1.69\%)} &
                        16.95 / 5.40 \textcolor{forestgreen}{(+0.99\%)}
                        \\
\multicolumn{1}{c|}{} & GPS-Net & \textbf{16.37} / \textbf{6.36}& \textbf{18.00} / \textbf{6.33} \textcolor{red}{(-0.03\%)} &
                        \underline{17.31} / \textbf{6.42} \textcolor{forestgreen}{(+0.06\%)}
                        \\ \hline
\multicolumn{1}{c|}{\multirow{4}{*}{R/mR@100}} 
                        & IMP & 13.06 / 2.68 & 15.32 / 4.38 \textcolor{forestgreen}{(+1.70\%)} &
                        15.63 / 5.57 \textcolor{forestgreen}{(\textbf{+2.89\%})}
                        \\
\multicolumn{1}{c|}{} & MOTIFS &  \underline{15.43} / \underline{4.65} & \underline{17.83} / \underline{6.38} \textcolor{forestgreen}{(+1.73\%)} &
                       \textbf{18.21} / \underline{6.14} \textcolor{forestgreen}{(+1.49\%)}
                        \\
\multicolumn{1}{c|}{} & VCTree & 14.62 / 4.54 &  17.58 / 6.30 \textcolor{forestgreen}{(\textbf{+1.76\%})} &
                        17.42 / 5.53 \textcolor{forestgreen}{(+0.99\%)}
                        \\
\multicolumn{1}{c|}{} & GPS-Net & \textbf{16.91} / \textbf{6.49} & \textbf{18.68} / \textbf{6.52} \textcolor{red}{(-0.03\%)} &
                        \underline{17.90} / \textbf{6.55} \textcolor{forestgreen}{(+0.06\%)}
                        \\ \Xhline{3\arrayrulewidth}
\end{tabular}%
}
\footnotesize{\\ 
$(\cdot)$ indicates the difference in mR@K when each algorithm is applied compared to FedAvg.
}
\end{table*}

Next, we verify whether the improvements in FL algorithms stay valid in our benchmark.
We conducted additional experiments on two FL algorithms employing momentum. Momentum-based update strategies prove effective in maintaining local training closer to the global update direction. By mitigating the adverse effects of data distribution discrepancies, these algorithms can enhance both convergence stability and model performance.

\subsubsection{FedAvgM}

We present the result of applying FedAvgM~\cite{fedavgm} in
\tabref{FedAvgM-Result}.
FedAvgM utilizes the momentum in updating a global model on the server side and relieves the varying directions of local updates due to the stochastic variance across clients.
  FedAvgM updates the global model as follows:
\begin{align}
    w_g^{r+1} &= w_g^r - v^r,\\
    v^r&=\beta v^{r-1} +\sum _{k=1}^K \frac{n_k}{n}\Delta w_k^r
\end{align}
where $\beta$ is the momentum hyperparameter for FedAvgM, 
$n_k$ is the number of examples,  
 $\Delta w_k^r$ is the weight update from $k$'s client, and $n=\sum _{k=1}^K n_k$.

\textbf{Results:} FedAvgM sufficiently improves the performance of all methods. 
For R/mR@20, R/mR@50, and R/mR@100, there are average performance improvements of +1.24\%, +1.27\%, and +1.30\% in the Shard-nonIID case, respectively.
These performance improvements under FedAvgM can be attributed to FedAvgM's momentum-based updates which help stabilize training by reducing local update oscillations in the optimization process, leading to faster convergence and better generalization across heterogeneous client data.
IMP, MOTIFS, and VCTree showed noticeable performance increases, while GPS-Net did not.
GPS-Net in the Shard non-IID case shows a negligible gap (i.e.,~$\leq 0.03\%$), indicating that GPS-Net already incorporates factors that mitigate the effects of heterogeneity.

\subsubsection{FedAdam}

FedOpt~\cite{fedopt} provides a framework for improving optimization, which
supports server-side optimization algorithms (e.g.,~Adam) to enhance
convergence and stability. This approach deals with divergent client data
distributions and fluctuations in client participation rates. FedOpt uses
different optimizers in local and global updates. In our case, we utilize the
Adam~\cite{kingma2014adam} optimizer for global updates:
\begin{align}
    m^{r+1}&=\beta_1 m^r+ (1-\beta_1)\Delta w^r,\\
      v^{r+1}&=\beta_2 v^r + (1-\beta_2)(\Delta w^r)^2,\\
      w_g^{r+1}&=w_g^r-\eta\frac{m^{r+1}}{\sqrt{v^{r+1}}+\epsilon},
\end{align}
where $\beta_1$ and $\beta_2$ are the momentum hyperparameters, $\epsilon$ is a small constant added to the denominator to ensure numerical stability and prevent division by $0$. We present the result of applying FedAdam in \tabref{FedAvgM-Result}.

\textbf{Results:} FedAdam demonstrated a marginally superior performance improvement compared to FedAvgM. For R/mR@20, R/mR@50, and R/mR@100, there are average performance improvements of {+}1.30\%, {+}1.35\%, and {+}1.36\% in the Shard-nonIID case, respectively. 
Interestingly, FedAdam shows a noticeable performance improvement when combined with IMP, as shown in the table. IMP has a lower initial performance (based on R/mR@20, R/mR@50, and R/mR@100) when compared to other methods (MOTIFS, VCTree, GPS-Net). However, when combined with FedAdam, it showed the greatest performance improvement ({+}2.89\% in R/mR@100 case).
IMP simply learns by iteratively updating relationships between objects. IMP is prone to learning by being overly head-class-biased in class imbalances, and performance degradation is inevitable in tail classes. In this environment, FedAdam has most likely improved its model effectively in the tail class, where losses are concentrated due to the long-tailed problem.
Contrary to IMP, GPS-Net has various strategies to solve the long-tailed problem.
Similar to the FedAvgM experimental result, GPS-Net showed no significant change in performance. 

We can conclude that the enhancement of FL algorithms is effective when dealing with scenarios involving diverse semantic information across clients. Furthermore, the experimental outcomes with IMP and GPS-Net reveal an intriguing connection: the long-tailed problem encountered in scene graph generation tasks shares notable similarities with the data heterogeneity issues faced in FL.
In scenarios where scene graph generation tasks must be addressed in a distributed data environment, selecting a scene graph generation method with a strong strategy for handling long-tailed problems or choosing an FL algorithm that effectively deals with data heterogeneity would significantly increase the likelihood of simultaneously tackling both challenges.

\begin{table*}[t!]
\centering
\caption{Comparisons of FL algorithms on CelebA emotion classification (Smiling vs. Not smiling) task}
  \begin{tabular}{c cc ccc}
  \Xhline{3\arrayrulewidth}
  \multirow{2}{*}{Method} & \multicolumn{2}{c}{Shard} & \multicolumn{3}{c}{Dirichlet distribution}\\
   & IID & non-IID & $\alpha=10$ & $\alpha=1$ & $\alpha=0.2$\\
    \hline
    FedAvg  & 90.94 (32) & 90.43 (52) & 91.56 (32) & 91.65 (37) & \underline{91.03} (47) \\
    FedAvgM & \textbf{92.93} (27) & \textbf{91.22} (32) & \textbf{93.4}  (27) & \textbf{92.93} (27) & \textbf{91.32} (37) \\
    FedAdam & \underline{91.84} (52) & \underline{91.03} (52) & \underline{91.71} (52) & \underline{91.77} (52) & 90.79 (52) \\
    \Xhline{3\arrayrulewidth}
  \end{tabular}
  \footnotesize{\\($\cdot$) is the communication rounds to reach 85\% Acc.}
  \label{tab:CelebA}
\end{table*}

\section{Additional Experiments on CelebA}
To demonstrate the generalizability, we also applied our clustering method to the CelebA dataset, which comprises 40 attributes (e.g.,~Eyeglasses, Wearing hat, Wavy hair), not a specific label.\footnote{We attach the details on CelebA experiments in the Appendix. \ref{Appendix: CelebA}}

\subsection{Construction of non-IID CelebA Dataset}
    In prior works, the CelebA dataset is distributed only according to identity in a non-IID case, e.g.,~each client has pictures of the same person, which limits the method for representing a kind of non-IID setting. It also cannot represent practical cases, such as the contents of CCTV or broadcast systems.
         Therefore, we build a non-IID dataset with the following steps.
        First, we perform K-Means clustering with 5 clusters. Since the attributes consist of binary values (-1, 1), we do not apply super-classes to preserve the meaning of each attribute.
        Second, we cluster images and allocate the cluster index to each image. Based on the cluster index of each image, we build non-IID cases by applying data partitioning methods such as shard-based and Dirichlet distribution-based partitions.
    We evaluate the classification task to check whether each person is smiling in \ref{tab:CelebA}.

    \subsection{Analysis}

    Similar to the result of the PSG task, it shows degradation of performance and a slow convergence rate in the shard non-IID case.
    For non-IID cases, including shard non-IID and Dirichlet $\alpha=0.2$, it shows a slower convergence rate than IID cases.
    For FedAvg, a non-IID environment, including shard non-IID and Dirichlet $\alpha=0.2$, leads to slower convergence than the IID cases, evidenced by slow convergence of 32$\rightarrow$52 under Shard and 32$\rightarrow$47 under Dirichlet-based partitions.
    For FedAvgM, it can also solve non-IID problems, where it reduces the communication cost by 20 more than FedAvg, especially in the Shard non-IID case 52$\rightarrow$32.
    For FedAdam, it shows the slowest convergence rate, conjecturing that it is because of the global learning rate of $1e^{-4}$.

\section{Conclusion}

Our work takes a decisive step toward closing the gap between federated learning (FL) research and high-level visual understanding by introducing the \emph{first} benchmark framework for multi-semantic vision tasks under controlled data heterogeneity.
Existing approaches for generating data heterogeneity rely on single-label datasets and cannot be extended to multiple semantic collections. To overcome this limitation, we propose a clustering-based scheme that groups samples based on the semantics of the images. This design naturally accommodates datasets with multiple labels per image, such as PSG and CelebA. 
We validate our framework on the PSG dataset and CelebA by partitioning the data according to cluster assignments and measuring convergence behavior.
As the data heterogeneity increases, i.e.,~from non-IID to IID, training converges more slowly and shows lower final accuracy for both datasets, confirming the ability to construct heterogeneity. Furthermore, our benchmark demonstrated consistent trends with prior FL studies, even when extended to various FL scenarios, i.e, changes in participation rates and number of clusters.

While the proposed approach shows significant results, it has limitations. First, determining the optimal number of clusters remains challenging. Second, the method assumes the availability of semantic annotations, which may not always be accessible.

 Nevertheless, the proposed benchmark can be adopted in practical use. It can be extended to multi-label and structured data tasks such as multimodal learning, visual question answering, and relation extraction. The framework is highly applicable to domains such as broadcasting and media, where centralized learning is often impractical due to stringent requirements regarding raw data confidentiality, intellectual property, and content ownership. FL with semantic-aware partitioning offers a collaborative model training without compromising sensitive media assets.

 \section*{Acknowledgments}
 This work was supported by
Institute of Information \&
Communications Technology Planning \& Evaluation (IITP) grant funded by the Korea government(MSIT)
(No.2021-0-00852, Development of Intelligent Media Attributes Extraction and Sharing Technology) and National Research Foundation of Korea (NRF) grant funded by the Korea government(MSIT) (No. RS-2024-00459023).

 \printcredits

\bibliographystyle{cas-model2-names}

\bibliography{cas-refs}
%\bibliography{cas-refs-Arxiv}

\clearpage
\appendix

\section{Detailed Desciption of the Dataset}\label{Detailed Desciption of Clustered Image}

\subsection{PSG dataset}

We utilize the PSG dataset \cite{yang2022panoptic} where each image is annotated with objects (including stuff such as ``sky" or ``grass"), panoptic segmentation masks, and fine-grained relationships between those objects.
PSG dataset leverages VG150 \cite{vg} and COCO \cite{lin2014microsoft} datasets
by integrating their comprehensive object and relationship annotations into the scene graph generation task. 
Specifically, the PSG dataset directly inherits the panoptic segmentation annotations from COCO. The VG150 dataset contains many `trivial' (not meaningful) predicates with the direction of predicates (e.g., of in hair-`of'-man, has in man-`has'-head), and the PSG dataset gets rid of these predicates.
The PSG dataset contains 133 objects and 374 relationships, sufficiently covering the diversity from the VG150 and COCO datasets while compensating for their limitations. These relationships are more detailed and extensive than in other datasets, allowing for richer scene graph representations.

\begin{table}[b!]
\centering
\caption{Dataset Information}
\label{Table: datainfo}
\resizebox{0.48\textwidth}{!}{
\begin{tabular}{c|cc}
\Xhline{3\arrayrulewidth}
\textbf{Data}    & \multicolumn{2}{c}{\textbf{Amount}}                         \\ \hline
PSG dataset - Train      & \multicolumn{2}{c}{46 K}                                    \\ \Xhline{3\arrayrulewidth}
\textbf{Cluster} & \multicolumn{1}{c|}{\textbf{Balanced (24.5\%)}} & \textbf{Imbalanced (100\%)} \\ \hline
Cluster 0        & \multicolumn{1}{c|}{2.2K (4.9\%)}        & 2.2K (4.9\%)              \\ \hline
Cluster 1        & \multicolumn{1}{c|}{2.2K (4.9\%)}        & 27K (58.1\%)              \\ \hline
Cluster 2        & \multicolumn{1}{c|}{2.2K (4.9\%)}       & 5.1K (10.9\%)              \\ \hline
Cluster 3        & \multicolumn{1}{c|}{2.2K (4.9\%)}       & 3.3K (7.1\%)              \\ \hline
Cluster 4        & \multicolumn{1}{c|}{2.2K (4.9\%)}       & 8.8K (19.0\%)              \\ \Xhline{3\arrayrulewidth}
\end{tabular}}
\end{table}

\subsection{Clustered PSG Dataset}\label{Clustered PSG Dataset}
\begin{figure*}[h!]
    \centering
\includegraphics[width=\linewidth]{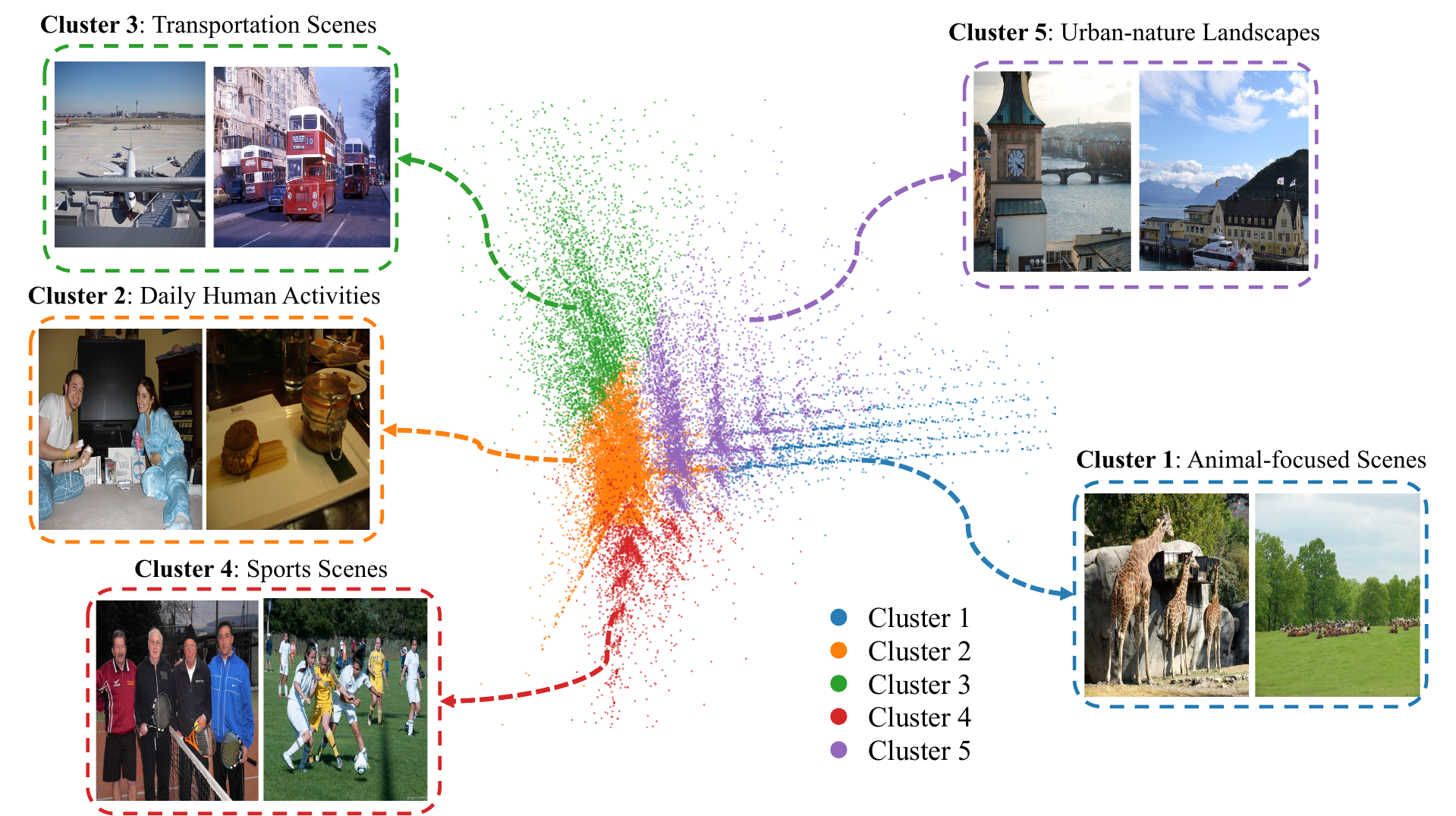}
    \caption{\textbf{Semantic clusters of PSG dataset} (a visualization via Principal Component Analysis)}
    \label{Fig:Example Images of Cluster}
\end{figure*}

Fig. \ref{Fig:Example Images of Cluster} illustrates the sample images from each cluster and the PCA clustering result image.
The qualitative visualization clearly demonstrates that our K-means Clustering of Category Tensor effectively and intuitively segments PSG dataset, leading to the splits given semantic information.

\section{Description of Baseline Algorithms}
\begin{figure*}[tb!]
\centering
\includegraphics[width=\linewidth]{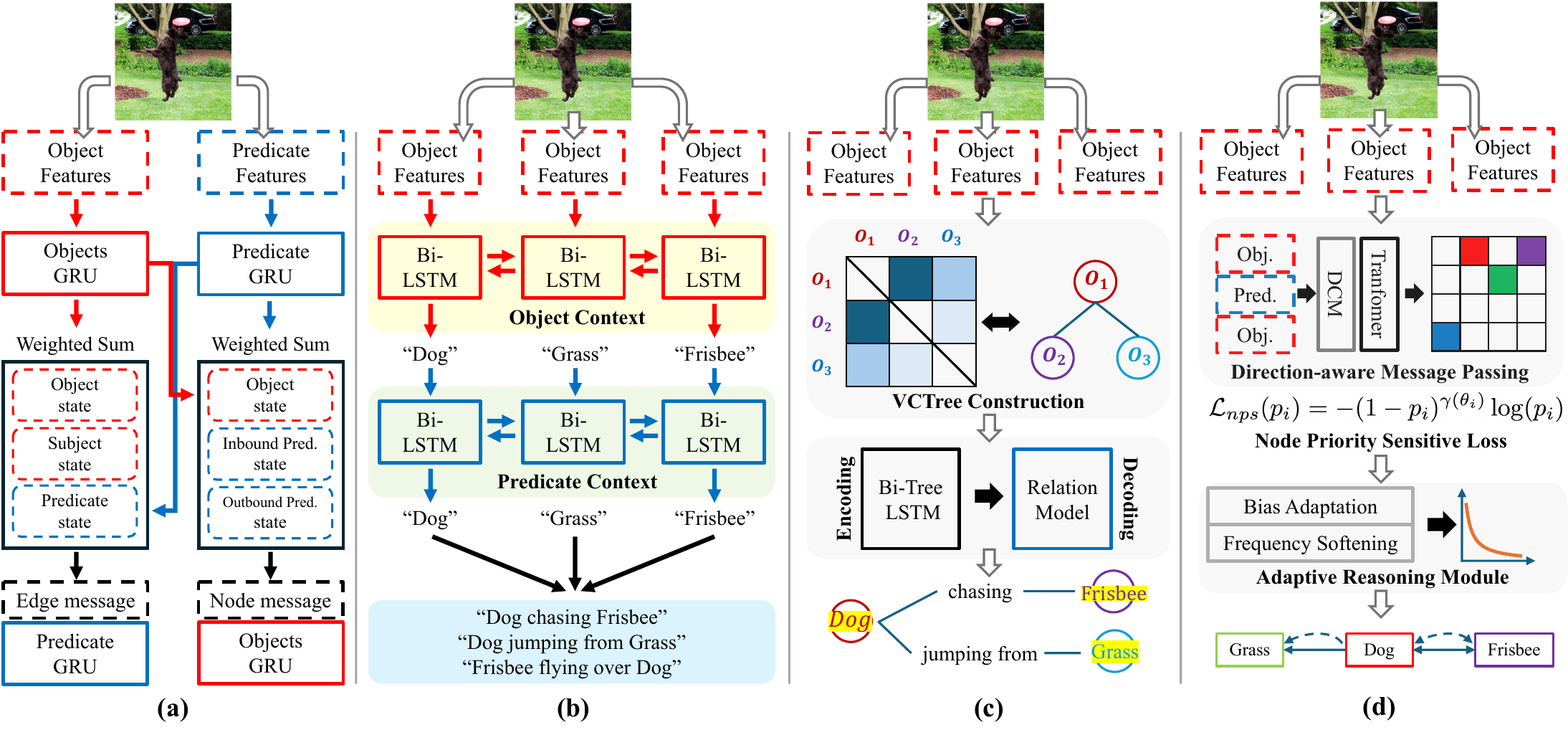}
\caption{\textbf{Overview of PSG Methods}
\textbf{(a)} IMP, 
\textbf{(b)} MOTIFS,
\textbf{(c)} VCTree,
\textbf{(d)} GPS-Net.
}
\label{Fig:psg_methods}
\end{figure*}

We choose four panoptic scene graph generation models to validate our benchmark. For ease of understanding, we present the overview of PSG methods in Fig. \ref{Fig:psg_methods}. \textbf{(a)} IMP\cite{imp} (Iterative Message Passing) is a method in scene graph generation that repeatedly delivers messages for each node (object) and edge (relationships) and updates information to optimize objects and relationships simultaneously with GRUs. \textbf{(b)} MOTIFS\cite{motifs} leverages the role of motifs in object relationships to guide scene graph generation, using bi-LSTMs to model context for better relationship prediction. \textbf{(c)} VCTree \cite{vctree} constructs a dynamic tree structure to hierarchically capture visual and semantic dependencies among objects using the bidirectional TreeLSTM, enabling effective representation of relationships for scene graphs. \textbf{(d)} GPS-Net\cite{gpsnet} enhances scene graph generation by integrating the Graph Property Sensing module that dynamically captures both global and local contextual properties of the graph. The Graph Property Sensing module consists of (1) Direction-aware Message Passing (DMP), (2) Node Priority Sensitive Loss (NPS-loss) utilizing object classification scores $p_i$ and (3) Adaptive Reasoning Module (ARM). It leverages an adaptive message-passing mechanism to encode object relationships better and address biases in the data, resulting in more accurate and context-aware scene graphs.

\section{Additional Experiments}\label{Additional Experiments}
\subsection{Convergence Behavior}
\begin{figure*}[tb!]
    \centering
    \includegraphics[width=0.32\linewidth]{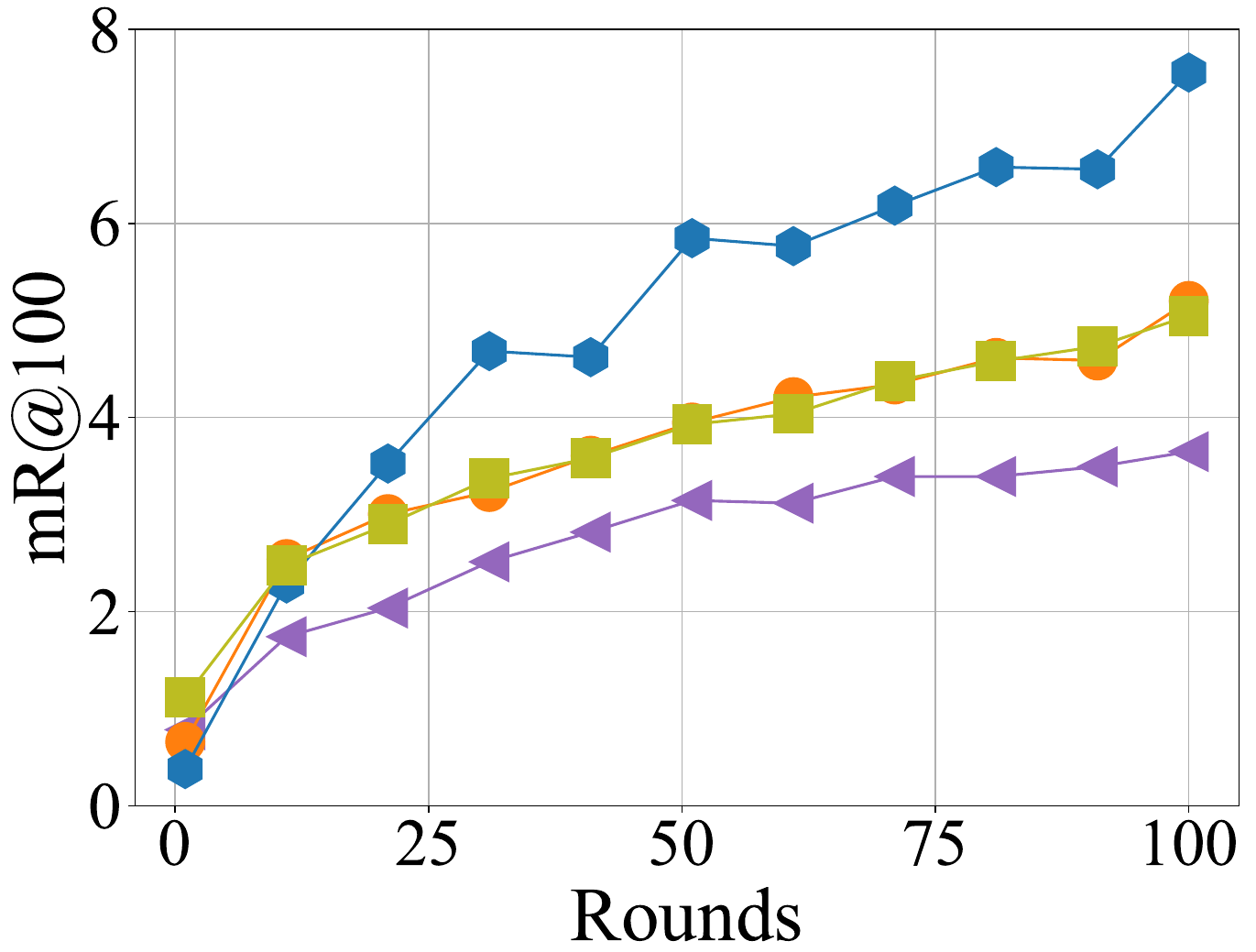}
    \hspace{0.05\linewidth}
    \includegraphics[width=0.32\linewidth]{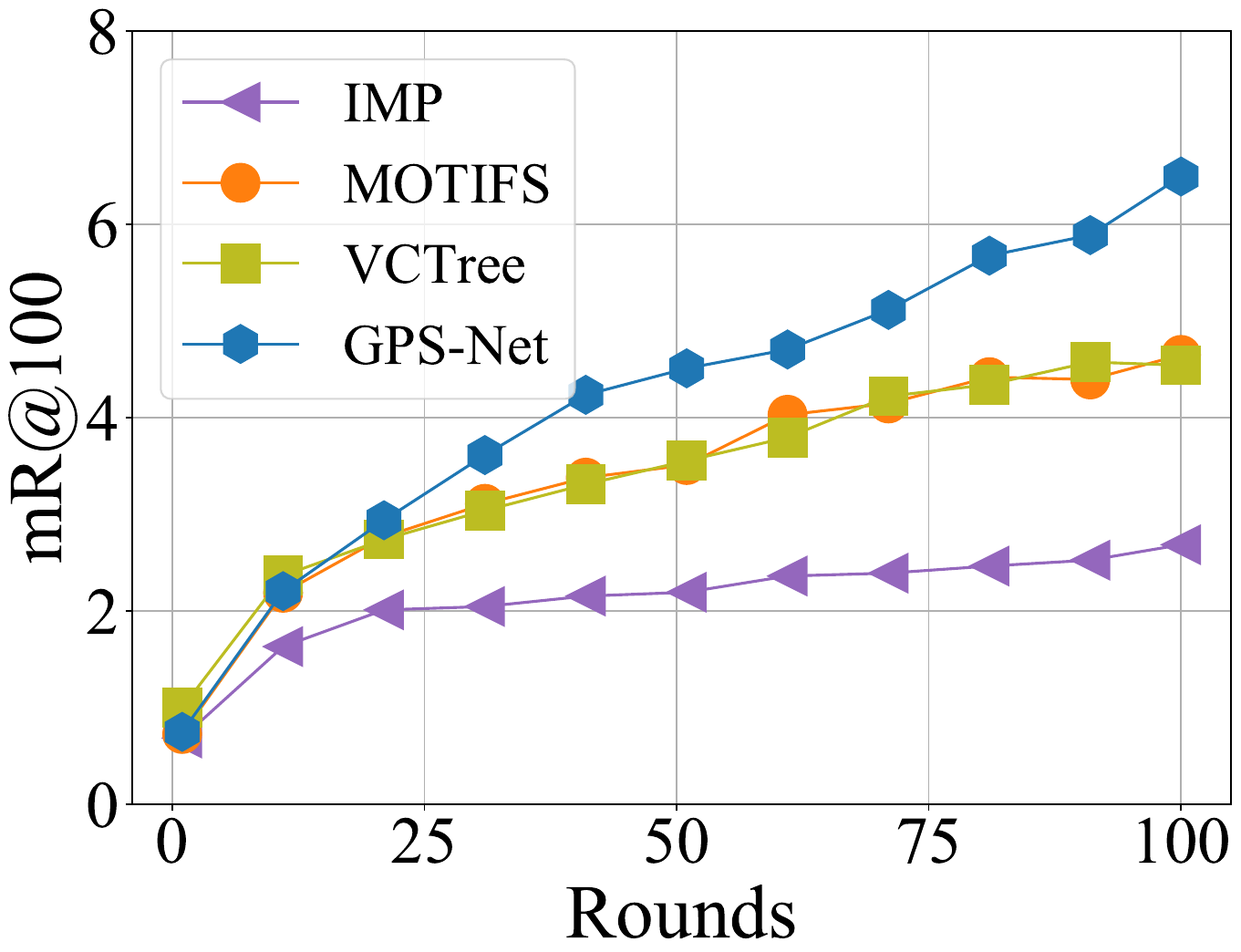}
\caption{Convergence behaviors of the balanced dataset for Shard IID (Left) and Shard non-IID (Right) cases}
\label{Fig:Convergence-Shard}
\end{figure*}

\begin{figure*}[tb!]
    \centering
    \includegraphics[width=0.32\linewidth]{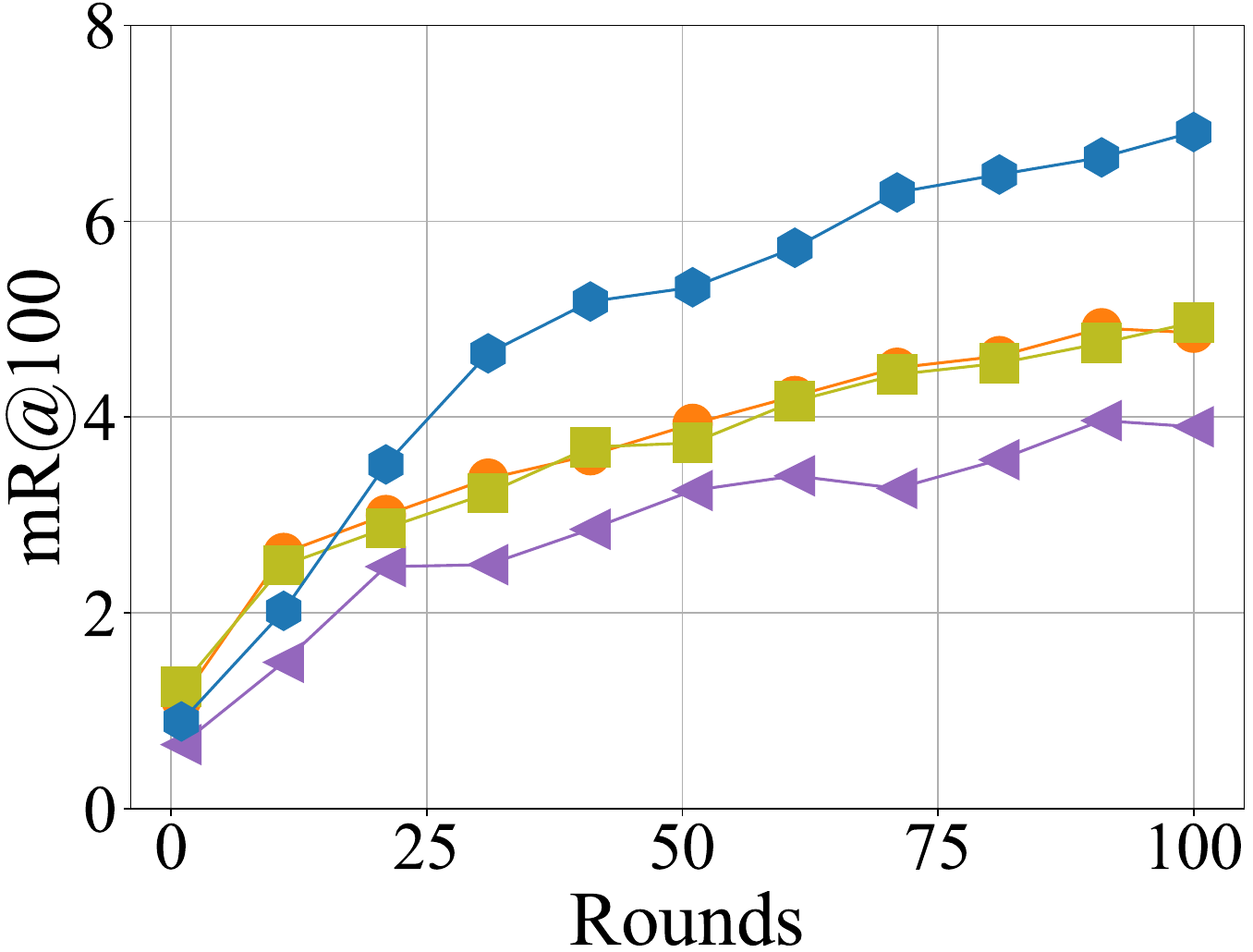}
    \includegraphics[width=0.32\linewidth]{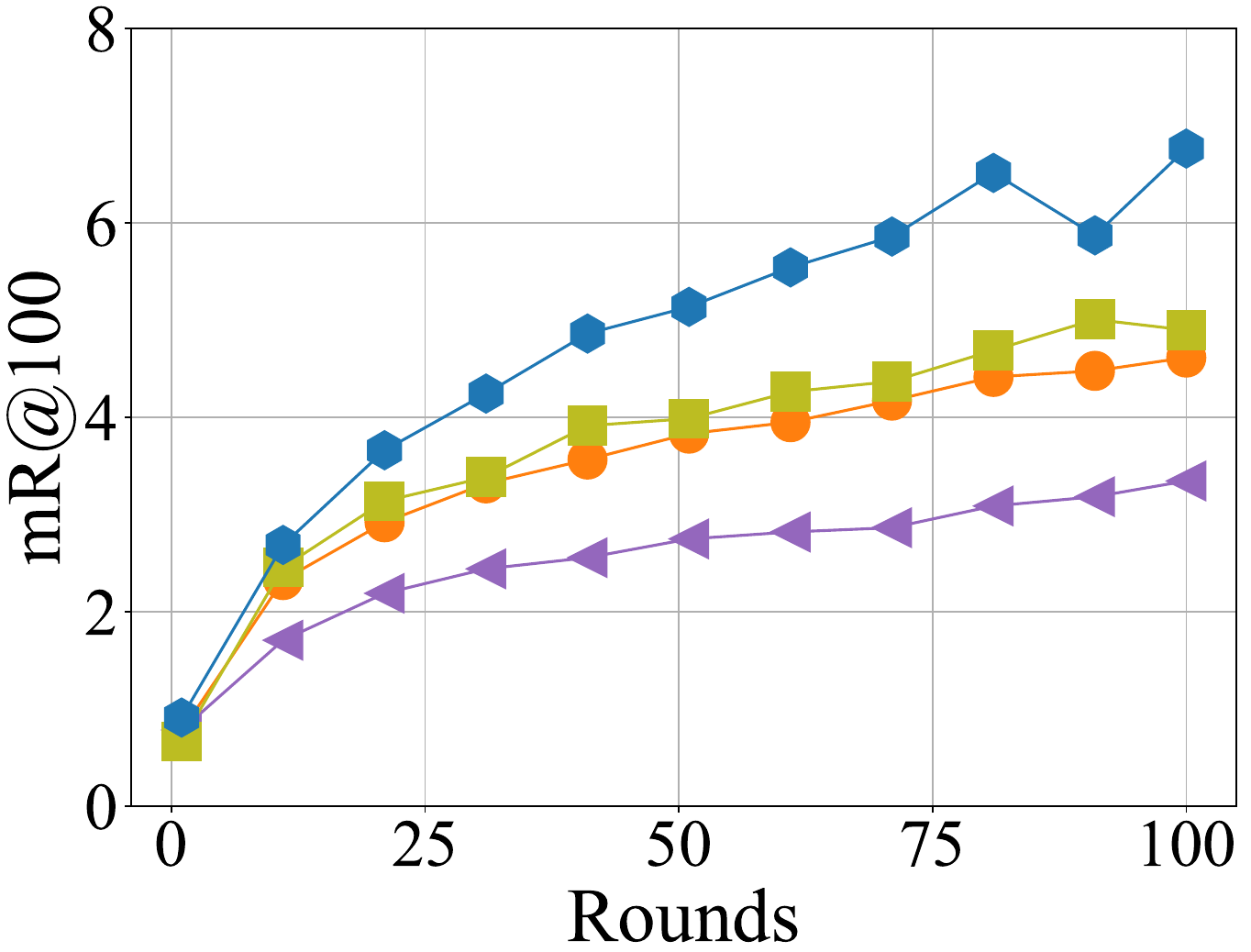}
    \includegraphics[width=0.32\linewidth]{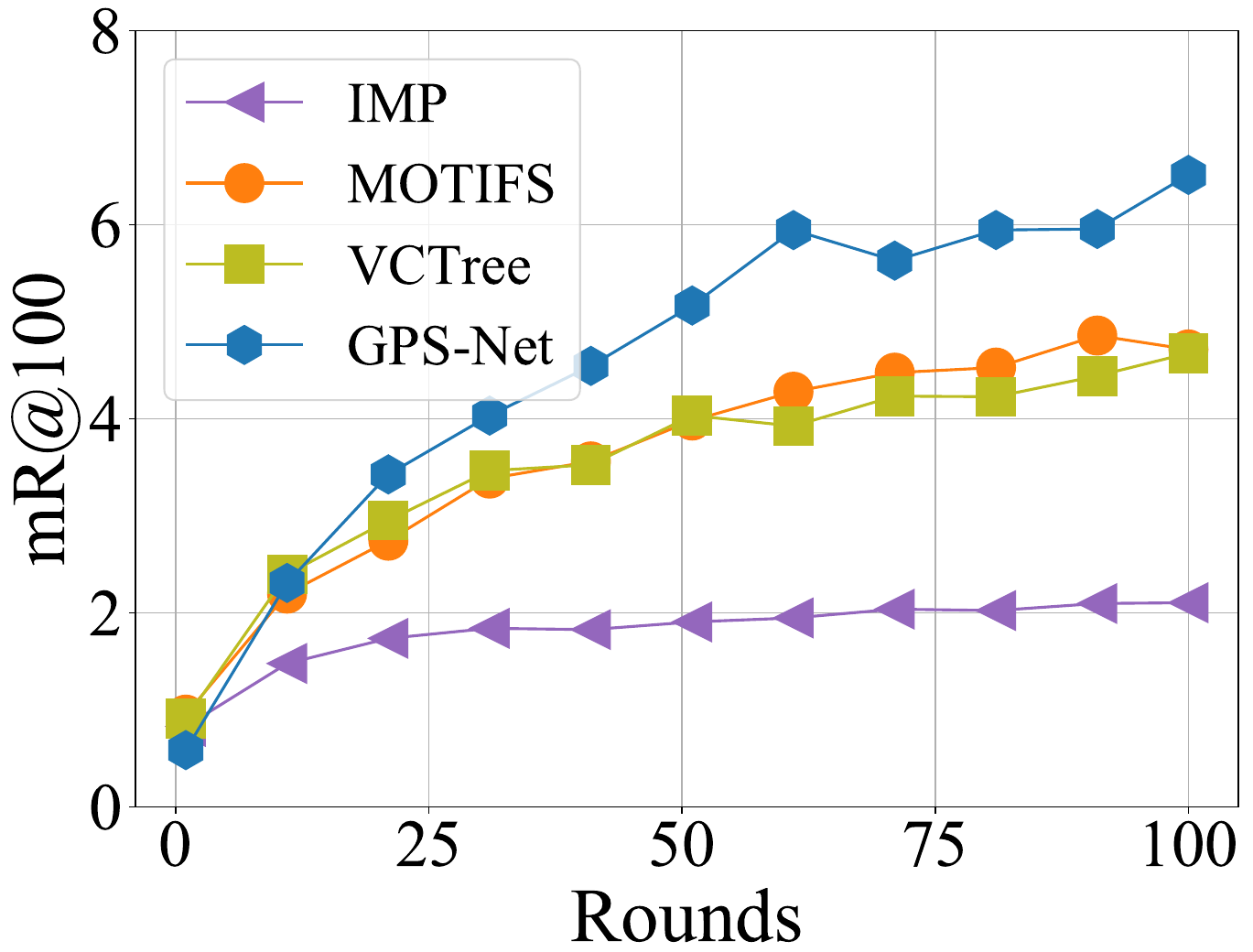}
\caption{Convergence behaviors of the balanced dataset for Dirichlet distribution (Left: $\alpha=10$, Middle: $\alpha=1$, Right: $\alpha=0.2$)}
\label{Fig:Convergence-Diri}
\end{figure*}

We present the convergence behaviors of four models on the shard and Dirichlet distribution-based partition method in Fig. \ref{Fig:Convergence-Shard}.
When we compare the non-IID and IID cases of shard, GPS-Net shows remarkable performance improvement in IID. GPS-Net has three key modules (DMP, NPS, ARM). In prior work \cite{yang2022panoptic}, DMP was the key module for the high performance in VG150, which has the direction of predicates in the dataset (e.g., of in hair-of-man, has in man-has-head). 
Because these predicates are removed in the PSG dataset, the DMP module has a lower effect on performance. However, we conjecture that other modules (NPS, ARM) designed to solve long-tailed problems are effective in the FL scenarios. As a result, the performance of GPS-Net shows the fastest convergence behavior.
Despite these modules, GPS-Net shows a decrease in convergence speed under the non-IID situation.
MOTIFS and VCTree show similar behaviors in IID and non-IID cases, and they also have the same model structures(LSTM) and do not consider the long-tailed problem. These two methods do not seem to be significantly affected in terms of convergence speed by the non-IID situation.
IMP also showed a critical decrease in convergence speed under the non-IID situation.

We also show the convergence behavior for Dirichlet distribution in Fig. \ref{Fig:Convergence-Diri}, where $\alpha$ is the $[10, 1, 0.2]$.
It shows similar results to behaviors of the shard-based partition scenario.
Overall, as the data heterogeneity increases, the convergence behaviors show decreased performance.
However, MOTIFS and VCTREE still do not seem to be significantly affected in terms of convergence speed in both Shard-based partition and Dirichlet distribution-based partition.

\subsection{Analysis of Communication Cost}
\begin{table}[tb!]
\centering
\caption{Comparison of the Communication Cost for Heterogeneity}
\label{Table: FL Communication Cost}
\resizebox{0.48\textwidth}{!}{
\begin{tabular}{cccc}
\Xhline{3\arrayrulewidth}
Method & \# of model parameters & Shard IID  & Shard non-IID \\ 
\Xhline{1.5\arrayrulewidth}
IMP & \textbf{32M} & 64(x 1)& 64(x 1)\\
MOTIFS & 63M & 63(x 0.98)& 63(x 0.98)\\
VCTree & 59M & 59(x 0.92)& 59(x 0.92)\\
GPS-Net & 37M & \textbf{37(x 0.57)}& \textbf{37(x 0.57)}\\
\Xhline{3\arrayrulewidth}
\end{tabular}}
\end{table}

Communication efficiency is one of the most important factors because communication costs are crucial in a practical FL scenario.
When thinking of communication rate, as in a scenario with limited communication resources, we compare the methods by rigorously measuring the actual communication costs required to reach the same level of performance.

From the convergence behavior in Fig. \ref{Fig:Convergence-Shard} and \ref{Fig:Convergence-Diri}, we compute the required communication costs to compare the communication efficiency of each method. Specifically, we calculate the total communication cost, i.e., the number of model parameters multiplied by the communication round, to reach the `mR@100 = 2’ (the reason for the target performance is that the IMP shows the worst performance and converges near 2).
We show the number of model parameters for each method and the resulting communication costs required in Table \ref{Table: FL Communication Cost}.

IMP has the smallest number of model parameters but the highest communication cost.
In contrast, GPS-Net has a similar number of model parameters with IMP, which accounts for half of the total communication cost, denoting that GPS-Net is the resource-efficient scene graph generation method in FL. GPS-Net shows remarkable performance because it has core startegies to resolve the long-tailed problem. We conjecture that it shows the rapid convergence behavior to higher accuracies with fewer communication costs.

\subsection{Cluster Imbalance Effect}

\begin{table*}[tb!]
\centering
\caption{Comparison of the performances on the Imbalanced Dataset}
\label{Table: FL-PSG data}

\begin{tabular}{cccccc}
\Xhline{3\arrayrulewidth}
\multirow{2}{*}{\shortstack{R/mR\\@K}}                                       & \multirow{2}{*}{Method} &                            &              & \multicolumn{2}{c}{Shard}   \\
                                                              &                             & CL\cite{yang2022panoptic}  & Random       & IID          & non-IID      \\ 
\Xhline{1.5\arrayrulewidth}
\multicolumn{1}{c|}{\multirow{4}{*}{\shortstack{R/mR\\@20}}}  & IMP                         & 16.5 / 6.52                & 16.10 / 5.68 & 16.38 / 5.97 & 15.50 / 4.75 \\
\multicolumn{1}{c|}{}                                         & MOTIFS                      & \underline{20.0} / \underline{9.10}                & \underline{16.66} / \underline{6.52} & 16.64 / 6.60 & 16.34 / \underline{6.32} \\
\multicolumn{1}{c|}{}                                         & VCTree                      & \textbf{20.6} / \textbf{9.70}                & 16.49 / 6.42 & \underline{16.93} / \underline{7.03} & \underline{16.46} / 6.09 \\
\multicolumn{1}{c|}{}                                         & GPS-Net                     & 17.8 / 7.03                & \textbf{17.90} / \textbf{7.29} & \textbf{18.12} / \textbf{7.38} & \textbf{17.98} / \textbf{8.10} \\
\Xhline{1.5\arrayrulewidth}
\multicolumn{1}{c|}{\multirow{4}{*}{\shortstack{R/mR\\@50}}}  & IMP                         & 18.2 / 7.05                & 17.53 / 6.10 & 17.74 / 6.43 & 16.89 / 5.11 \\
\multicolumn{1}{c|}{}                                         & MOTIFS                      & \underline{21.7} / \underline{9.57}                & \underline{18.26} / \underline{7.03} & 18.38 / 7.12 & 17.89 / \underline{6.70} \\
\multicolumn{1}{c|}{}                                         & VCTree                      & \textbf{22.1} / \textbf{10.2}                & 17.94 / 6.84 & \underline{18.47} / \underline{7.44} & \underline{17.97} / 6.52 \\
\multicolumn{1}{c|}{}                                         & GPS-Net                     & 19.6 / 7.49                & \textbf{19.44} / \textbf{7.77} & \textbf{19.78} / \textbf{7.85} & \textbf{19.36} / \textbf{8.46} \\
\Xhline{1.5\arrayrulewidth}
\multicolumn{1}{c|}{\multirow{4}{*}{\shortstack{R/mR\\@100}}} & IMP                         & 18.6 / 7.23                & 18.09 / 6.28 & 18.20 / 6.58 & 17.26 / 5.22 \\
\multicolumn{1}{c|}{}                                         & MOTIFS                      & \underline{22.0} / \underline{9.69}                & \underline{18.75} / \underline{7.23} & 18.88 / 7.23 & 18.39 / \underline{6.83} \\
\multicolumn{1}{c|}{}                                         & VCTree                      & \textbf{22.5} / \textbf{10.2}                & 18.52 / 7.00 & \underline{18.92} / \underline{7.56} & \underline{18.47} / 6.64 \\
\multicolumn{1}{c|}{}                                         & GPS-Net                     & 20.1 / 7.67                & \textbf{19.89} / \textbf{7.86} & \textbf{20.14} / \textbf{7.94} & \textbf{19.83} / \textbf{8.62} \\
\Xhline{3\arrayrulewidth}
\end{tabular}
\end{table*}

\textbf{Benchmark setups:}
To observe the effects of cluster imbalance, we do not equalize the number of data points.
The rationale for cluster balancing is that the imbalance across clusters probably prevents rigorous evaluations of FL’s capability to handle semantic heterogeneity because a model becomes overfitted to dominant clusters without balanced training across different semantics. 
Consequently, we believe that the balancing step is required to get rid of the overfitting issue.
    Also, we want to emphasize that the balanced setting, even with a quite smaller size, our result in \ref{Table: FL-PSG data} show sufficient and acceptable training behavior, i.e., shows regular convergence behavior and the discriminated final performances.

We tested 3 types of data partitioning as follows:

\textbf{(1) Random:} data is distributed randomly among all clients, ensuring nearly equal sizes for each. 

\textbf{(2) Shard-based partition IID:} We set $p=5$, where $p$ is the number of clusters that client sample from. As aforementioned, when $p$ equals the number of clusters, the data from each cluster is equally distributed among 100 clients.

\textbf{(3) Shard-based partition non-IID:} We set $p=1$ for imposing semantic heterogeneity. This time, clients are allocated to each cluster based on the amount of data in each cluster rather than assigning an equal number of clients to all clusters. 

\textbf{Results:} The results in Table \ref{Table: FL-PSG data} differ in some aspects from the analysis in the main paper. 
Firstly, the performance of all methods in FL scenarios has significantly increased. This is because the quantity of data assigned to each client has greatly increased, and models are overfitted to the dominant cluster. The performance trends of each method have changed as follows: In the case of IMP, the performance gap between CL and IID has significantly narrowed. While the performance in FL scenarios has greatly improved, the performance of CL has remained unchanged. This indicates that IMP is relatively less affected by the amount of data. In the case of MOTIFS and VCTree, the performance trends were almost similar to those observed in the previous experiments on the balanced dataset.
VCTree still appears to be slightly more vulnerable to data heterogeneity compared to MOTIFS. These methods, i.e., IMP, MOTIFS, and VCTree, show that performance decreases as data heterogeneity increases. In contrast, the trends observed for GPS-Net were completely different from what was expected generally in FL scenarios. The performance reported in the previous study \cite{yang2022panoptic} is lower than the performance on the balanced dataset. Prior research mentioned that the key point of GPS-Net explicitly models the direction of predicates, which is why it does not perform well on PSG dataset. However, when examining the results in our main table, modeling the direction of predicates might not be the cause.
Furthermore, GPS-Net showed the best results in the non-IID scenario. This is unusual and significantly deviated from our expectations. Therefore, we concluded that GPS-Net performs well when clients have sufficient data and a relatively small number of categories. Additionally, we believe that a detailed performance comparison through FL on the SGG task, rather than PSG task, would allow for a more in-depth analysis.

\subsection{Various FL Scenarios}

\begin{table}[t!]
    \centering
        \caption{Comparison of performances for the number of total clients.}
        \label{Table: Effect Of Total Client}
        \begin{tabular}{cccc}
            \Xhline{3\arrayrulewidth}
            R/mR@100 & Method & Shard IID & Shard non-IID \\ \Xhline{1.5\arrayrulewidth}
            \multicolumn{1}{c|}{\multirow{4}{*}{Clients 50}} 
                                  & IMP & 15.62 /  4.60 & 14.92 /  4.70 \\
            \multicolumn{1}{c|}{} & MOTIFS & \underline{18.24} / \textbf{7.74} & \textbf{18.22} / \underline{6.88}\\
            \multicolumn{1}{c|}{} & VCTree & 16.61 / 6.51 & 16.89 / 6.40 \\
            \multicolumn{1}{c|}{} & GPS-Net & \textbf{18.54} / \underline{7.22} & \underline{18.11} / \textbf{7.29} \\ \Xhline{1.5\arrayrulewidth}
            \multicolumn{1}{c|}{\multirow{4}{*}{Clients 100}} 
                                  & IMP & 14.45 / 3.65 & 13.06 / 2.68 \\
            \multicolumn{1}{c|}{} & MOTIFS & \underline{15.38} / \underline{5.20} & \underline{15.43} / \underline{4.65} \\
            \multicolumn{1}{c|}{} & VCTree & 14.97 / 5.05 &	14.62 / 4.54 \\
            \multicolumn{1}{c|}{} & GPS-Net & \textbf{17.08} / \textbf{7.55} & \textbf{16.91} /\textbf{ 6.49} \\ \Xhline{1.5\arrayrulewidth}
            \multicolumn{1}{c|}{\multirow{4}{*}{Clients 200}} 
                                  & IMP & 12.80 / 3.00 & 12.37 / 2.30 \\
            \multicolumn{1}{c|}{} & MOTIFS & \textbf{15.22} / \textbf{5.23} & \textbf{15.43} / \underline{5.08} \\
            \multicolumn{1}{c|}{} & VCTree & 12.54 / 3.56 & 12.4 / 3.33 \\
            \multicolumn{1}{c|}{} & GPS-Net & \underline{13.69} / \underline{5.22} & \underline{13.49} / \textbf{5.09} \\ \Xhline{3\arrayrulewidth}
        \end{tabular}
    \end{table}

In this section, we conduct a series of experiments to investigate the impact of different factors on federated learning performance. Federated learning operates in diverse environments, making it essential to test various scenarios to better understand how the approach performs. By manipulating key parameters such as the total number of clients, participation rates, and federated learning algorithms, we provide a comprehensive analysis of their effects on overall performance.

\textbf{Total clients:}
We evaluate performances to examine the effect of number of total clients, i.e., 50, 100, and 200, in Table \ref{Table: Effect Of Total Client}. 
For the case of 50 clients, each client has twice as much data per client compared to the 100 clients case.
Performance improves as the number of data samples that the clients have increased. 
Notably, VCTree shows a $2\times$ larger mR@K for 50 clients compared to 200 clients in the Shard non-IID case, indicating that VCTree is highly sensitive to the number of data samples, a critical factor in FL settings.
In contrast, MOTIFS and GPS-Net are less affected by the number of data samples, with GPS-Net achieving an mR@K greater than 5.09 and MOTIFS exceeding 4.65, significantly outperforming both IMP and VCTree across all cases.

\textbf{Participation rates:}
\begin{table}[t!]
        \centering
        \caption{Comparison of performances for participation rates}
        \label{Table: Effect Of Participate Client}
        \resizebox{0.48\textwidth}{!}{
        \begin{tabular}{cccc}
            \Xhline{3\arrayrulewidth}
            R/mR@100 & Method & Shard IID & Shard non-IID \\ \Xhline{1.5\arrayrulewidth}
            \multicolumn{1}{c|}{\multirow{4}{*}{\# of clients 5}} 
                                  & IMP & 14.45 / 3.65 & 13.06 / 2.68 \\
            \multicolumn{1}{c|}{} & MOTIFS & \underline{15.38} / \underline{5.20} & \underline{15.43} / \underline{4.65} \\
            \multicolumn{1}{c|}{} & VCTree & 14.97 / 5.05 & 14.62 / 4.54 \\
            \multicolumn{1}{c|}{} & GPS-Net & \textbf{17.08} /\textbf{ 7.55} & \textbf{16.91} / \textbf{6.49} \\ \Xhline{1.5\arrayrulewidth}

            \multicolumn{1}{c|}{\multirow{4}{*}{\# of clients 20}}
                                & IMP & 13.91 / 3.25 & 15.31 / 4.04 \\
            \multicolumn{1}{c|}{} & MOTIFS & \textbf{17.3} /\textbf{ 6.39} & \textbf{16.83} / \textbf{6.23} \\
            \multicolumn{1}{c|}{} & VCTree & 15.03 / 4.83 & 14.92 / 4.73  \\
            \multicolumn{1}{c|}{} & GPS-Net & \underline{16.81} / \underline{6.36} & \underline{16.66} / \underline{6.04} \\ \Xhline{3\arrayrulewidth}
        \end{tabular}}
\end{table}
We also evaluate the performance according to the number of participating clients, i.e., 5 and 20, in Table \ref{Table: Effect Of Participate Client}. 
As the number of participants increased, the performance of MOTIFS improved remarkably. 
According to previous FL studies, mainly handled the image classification task, increasing participation rate leads to improved performance in FL environments. However, in the PSG task, as the number of users increases, the performance does not show similar behaviors without MOTIFS. 
In other words, rather than increasing the number of participants in each round, a larger amount of data for each participant can result in greater performance improvement in this task.

Additionally, we have to focus on the performances of IMP in various FL scenarios. IMP is the oldest method in our experiments and shows a lower performance in the experiments of existing studies \cite{vctree, yang2022panoptic}. Therefore, the performance seems poor before being affected by data heterogeneity, making a detailed comparison difficult.

\subsection{Ablation Study on Number of Clusters}

\begin{table*}[t!]
    \small
    \centering
    \caption{Ablation studies on the number of clusters of the proposed FL benchmark.}
    \label{tab:K-ablation}
    \begin{tabular}{cccccccc}
    \Xhline{3\arrayrulewidth}
    \multirow{2}{*}{\shortstack{R/mR\\@K}}                        & \multirow{2}{*}{Method} & \multicolumn{2}{c}{n = 3}   & \multicolumn{2}{c}{n = 5}  & \multicolumn{2}{c}{n = 10}              \\
                                                                  &                            & IID          & non-IID      & IID & non-IID   & IID & non-IID \\
    \Xhline{1.5\arrayrulewidth}
    \multicolumn{1}{c|}{\multirow{4}{*}{\shortstack{R/mR\\@20}}}  & IMP     & 14.34 / 4.26 & 10.91 / 2.03 & 12.62 / 3.20 & 11.26 / 2.28 & 10.23 / 1.93 & 11.00 / 2.38 \\
    \multicolumn{1}{c|}{}                                         & MOTIFS  & \underline{14.41} / 4.72 & 13.90 / 3.85 & \underline{13.26} / \underline{4.64} & \underline{13.33} / \underline{4.06} & 11.02 / 3.07 & 10.82 / 2.98 \\
    \multicolumn{1}{c|}{}                                         & VCTree  & 14.08 / \underline{4.73} & \underline{14.37} / \underline{4.72} & 13.00 / 4.57 & 12.49 / 3.99 & \underline{11.17} / \underline{3.21} & \underline{11.29} / \underline{3.02} \\
    \multicolumn{1}{c|}{}                                         & GPS-Net & \textbf{16.43} / \textbf{6.71} & \textbf{15.83} / \textbf{6.05} & \textbf{14.83} / \textbf{6.90} & \textbf{14.57} / \textbf{5.90} & \textbf{12.85} / \textbf{4.93} & \textbf{12.55} / \textbf{4.28} \\
    \Xhline{1.5\arrayrulewidth}
    \multicolumn{1}{c|}{\multirow{4}{*}{\shortstack{R/mR\\@50}}}  & IMP     & 15.77 / 4.65 & 12.30 / 2.33 & 13.97 / 3.53 & 12.57 / 2.59 & 11.53 / 2.18 & 12.36 / 2.66 \\
    \multicolumn{1}{c|}{}                                         & MOTIFS  & \underline{15.97} / \underline{5.14} & 15.48 / 4.24 & \underline{14.82} / \underline{5.06} & \underline{14.92} / \underline{4.48} & 12.45 / 3.41 & 12.23 / \underline{3.39} \\
    \multicolumn{1}{c|}{}                                         & VCTree  & 15.56 / 5.09 & \underline{15.90} / \underline{5.05} & 14.50 / 4.94 & 14.04 / 4.41 & \underline{12.68} / \underline{3.61} & \underline{12.63} / 3.37 \\
    \multicolumn{1}{c|}{}                                         & GPS-Net & \textbf{17.85} / \textbf{7.08} & \textbf{17.36} / \textbf{6.41} & \textbf{16.42} / \textbf{7.37} & \textbf{16.37} / \textbf{6.36} & \textbf{14.29} / \textbf{5.28} & \textbf{14.04} / \textbf{4.70}\\
    \Xhline{1.5\arrayrulewidth}
    \multicolumn{1}{c|}{\multirow{4}{*}{\shortstack{R/mR\\@100}}} & IMP     & 16.27 / 4.76 & 12.83 / 2.44 & 14.45 / 3.65 & 13.06 / 2.68 & 12.05 / 2.30 & 12.90 / 2.78 \\
    \multicolumn{1}{c|}{}                                         & MOTIFS  & \underline{16.48} / \underline{5.26} & 16.03 / 4.40 & \underline{15.38} / \underline{5.20} & \underline{15.43} / \underline{4.65} & 13.02 / 3.58 & 12.92 / \underline{3.58}  \\
    \multicolumn{1}{c|}{}                                         & VCTree  & 16.01 / 5.20 & \underline{16.37} / \underline{5.20} & 14.97 / 5.05 & 14.62 / 4.54 & \underline{13.21} / \underline{3.77} & \underline{13.22} / 3.54 \\
    \multicolumn{1}{c|}{}                                         & GPS-Net & \textbf{18.31} / \textbf{7.19} & \textbf{17.80} / \textbf{6.51} & \textbf{17.08} / \textbf{7.55} & \textbf{16.91} / \textbf{6.49} & \textbf{14.85} / \textbf{5.41} & \textbf{14.66} / \textbf{4.83} \\
    \Xhline{3\arrayrulewidth}
    \end{tabular}
\end{table*}

We have repeated to test our partitioning pipeline on the PSG dataset using different number of clusters, e.g., $n=3$ and $n=10$ semantic clusters, and conducted the federated learning. We present the results on Table \ref{tab:K-ablation}. In both cases the qualitative ordering of PSG algorithms—GPS-Net $>$ MOTIFS $\approx$ VCTree $>$ IMP—remains unchanged, and the IID $\leftrightarrow$ non-IID performance gap scales smoothly with the strength of the induced heterogeneity. However, very fine clustering ($n=10$) exaggerates data-scarcity effects and masks the heterogeneity gap. These outcomes confirm that, regardless of the exact number of clusters used and or how sharply they are separated, our pipeline continues to provide a valid data splits of federated learning with a semantic heterogeneity.

\section {Discussion: Why FL Benchmarks for Scene Understanding is Required}

In this section, we discuss the necessity and potential extensions of our benchmarks.
In the media industry, companies are increasingly focused on protecting the copyrights of their original content. As a result, they have been reluctant to share raw media data externally. At the same time, there is a growing demand for leveraging AI in the media production process. Given this context, there is a pressing need for a model that can effectively learn from media content while preserving copyright. To fulfill this need, we proposed the FL scenario that allows for AI training without requiring the exposure of raw data. We believe this approach aligns well with the concept of protecting media content copyrights. Among the various tasks that utilize publicly available benchmark datasets, PSG plays a crucial role in understanding the context of media content. Scene graphs represent the relationships between objects, thus extracting the semantic meaning of the content.

This study offers a new direction for FL research, demonstrating its potential to support complex semantic learning tasks while preserving data privacy. The findings are particularly relevant for applications in sensitive domains such as healthcare and media, where privacy concerns limit centralized data sharing. Future work could extend the proposed benchmark to diverse vision and multimodal datasets, further enhancing its generalization and utility.

\begin{figure*}[h!]
    \centering
    \includegraphics[width=0.92\linewidth]{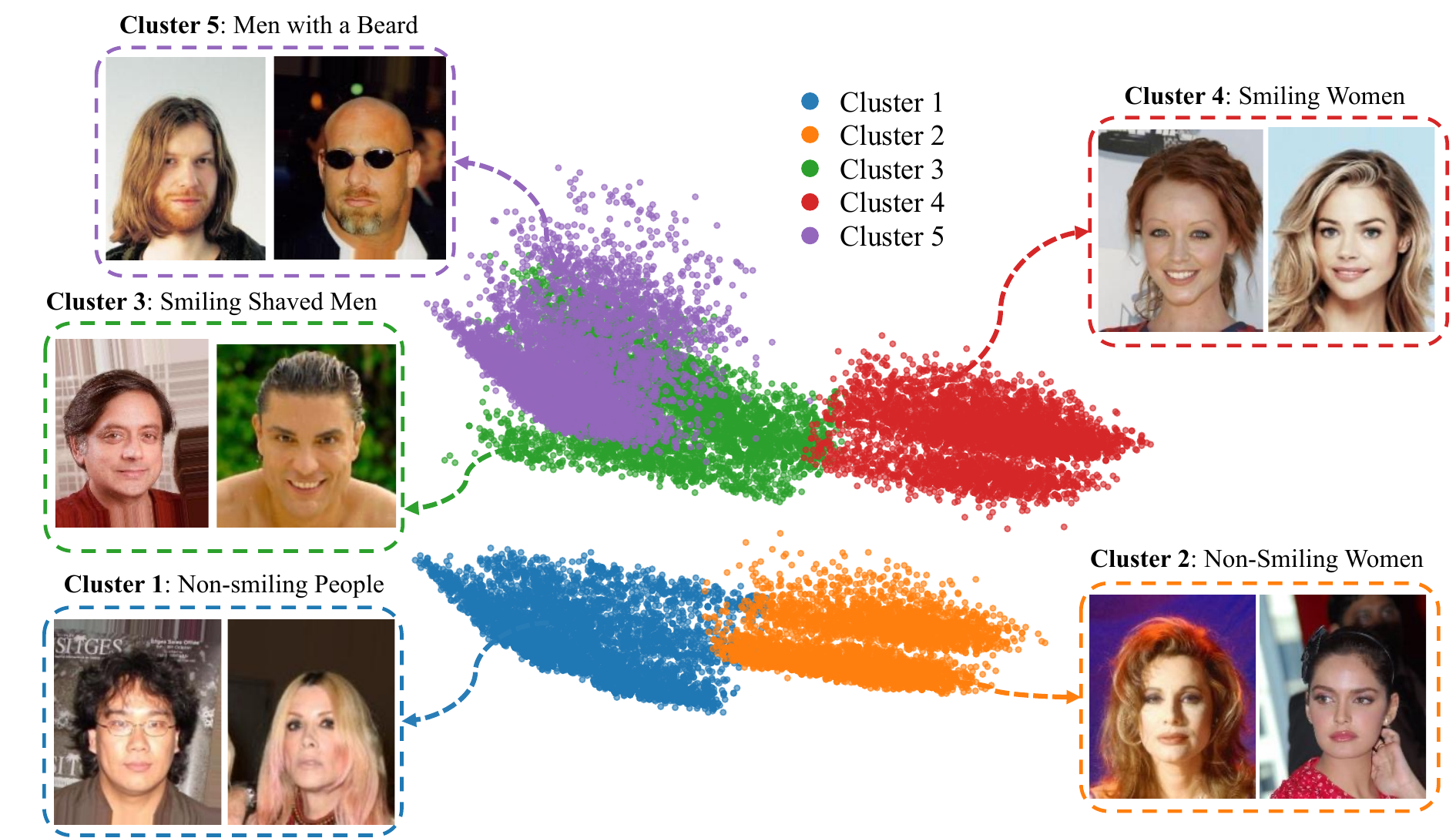}
    \caption{Semantic clusters of CelebA dataset}
    \label{fig:CelebA-cluster}
\end{figure*}

\section{Details on CelebA Experiments}\label{Appendix: CelebA}

\subsection{Clustered CelebA Dataset}
 We also applied our clustering method to the CelebA dataset, which is composed of 40 attributes (e.g., Eyeglasses, Wearing hat, Wavy hair, etc) and does not have a specific label for each image. Therefore, we build a non-IID dataset with the following steps.
First, we perform K-Means clustering with 5 clusters. Since the attributes consist of binary values (-1, 1), we do not apply super-classes to preserve the meaning of each attribute.
Second, we cluster images and allocate the cluster index to each image. Based on the cluster index of each image, we build non-IID cases by applying data partitioning methods such as shard-based and Dirichlet distribution-based partitions.
Figure \ref{fig:CelebA-cluster} illustrates the PCA clustering result. The qualitative visualization clearly demonstrates that our K-means Clustering leads to the splits given the semantic information.
\subsection{CelebA in FL}
We evaluate the classification task to check whether each person is smiling in Table \ref{tab:CelebA}. Similar to the result of the PSG task, it shows degradation of performance and a slow convergence rate in the shard non-IID case.
In prior works, the CelebA dataset is distributed only according to identity in a non-IID case, e.g., each client has pictures of the same person, which limits the method for representing a kind of non-IID setting.
Additionally, it can't represent non-IID cases, such as CCTV, or the behavior of small devices, because all identities of each client are the same. Our clustering method builds a more practical environment, not a fixed one.
For the hyperparameter, we set the local learning rate as $0.01$, momentum ratio for FedAvgM as $0.7$, and global learning rate as $1e^{-4}$ for FedAdam in $[1e^{-1}, 1e^{-2}, 1e^{-3}, 1e^{-4}]$.
The total number of rounds is $100$, and the participation ratio is $5\%$ in total number of clients $100$.

\end{document}